\pdfoutput=1

\documentclass[11pt]{article}

\usepackage[final]{coling}

\usepackage{times}
\usepackage{latexsym}

\usepackage[T1]{fontenc}

\usepackage[utf8]{inputenc}

\usepackage{microtype}

\usepackage{inconsolata}

\usepackage{graphicx}
\usepackage{booktabs}
\usepackage{mwe}
\usepackage{multirow}%
\usepackage{amsmath,amssymb,amsfonts}%
\usepackage{mathrsfs}%
\usepackage[title]{appendix}%
\usepackage{textcomp}%
\usepackage{manyfoot}%
\usepackage{algorithm}%
\usepackage{algorithmicx}%
\usepackage{algpseudocode}%
\usepackage{listings}%
\usepackage{times}
\usepackage{latexsym}
\usepackage{paralist}
\usepackage[utf8]{inputenc}
\usepackage{microtype}
\usepackage{multirow}
\usepackage{caption}
\usepackage{subcaption}
\usepackage[rightcaption]{sidecap}
\sidecaptionvpos{table}{c}
\sidecaptionvpos{figure}{c}
\usepackage{wrapfig,lipsum}
\usepackage{mdframed}
\usepackage{hyperref}
\usepackage{tcolorbox}
\usepackage[normalem]{ulem}
\usepackage{lipsum}
\usepackage{array}
\usepackage{makecell}
\usepackage{cleveref}
\usepackage{paralist}
\usepackage{soul}
\usepackage{xcolor}
\usepackage{colortbl}
\usepackage{float}
\definecolor{ForestGreen}{RGB}{55,190,55}

\newcommand{\rvs}[1]{\textsc{REVerSum}}

\title{\rvs{}: A Multi-staged Retrieval-Augmented Generation Method to Enhance Wikipedia Tail Biographies through Personal Narratives}

\author{Sayantan Adak, Pauras Mangesh Meher, Paramita Das \and Animesh Mukherjee\\
  IIT, Kharagpur \\
  West Bengal -- 721302
}

\begin{document}
\maketitle
\begin{abstract}
Wikipedia is an invaluable resource for factual information about a wide range of entities. However, the quality of articles on less-known entities often lags behind that of the well-known ones. This study proposes a novel approach to enhancing Wikipedia's B and C category biography articles by leveraging personal narratives such as autobiographies and biographies. By utilizing a multi-staged retrieval-augmented generation technique -- \rvs{} -- we aim to enrich the informational content of these lesser-known articles. Our study reveals that personal narratives can significantly improve the quality of Wikipedia articles, providing a rich source of reliable information that has been underutilized in previous studies. Based on crowd-based evaluation, \rvs{} generated content outperforms the best performing baseline by 17\% in terms of integrability to the original Wikipedia article and 28.5\% in terms of informativeness. \footnote{Code and Data are available at \url{https://github.com/sayantan11995/wikipedia_enrichment}}

\end{abstract}

\section{Introduction}
 Wikipedia plays a pivotal role in many areas of natural language processing (NLP) research, serving as a rich resource for pre-training machine learning models, fact verification, and as an external knowledge base. For instance, \citet{touvron2023llamaopenefficientfoundation}, \citet{thoppilan2022lamda}, and \citet{brown2020language} incorporate Wikipedia in their pre-training corpora. \citet{chen-etal-2017-reading} utilize Wikipedia to answer open-domain questions, while \citet{kirchenbauer2024hallucination} leverage it in a retrieval augmented generation (RAG) setup to reduce hallucination in question answering. In addition, \citet{reid2022can} use Wikipedia as an external resource to improve offline reinforcement learning tasks. However, in spite of its extensive usage and popularity, several categories on Wikipedia either lack decent coverage or the articles are not of acceptable quality. Creating new articles and editing older ones consumes significant time and resources, making it an expensive endeavor~\cite{10.1145/2682571.2797073}. 
Despite advances in text generation and retrieval-based modeling architectures, the automatic creation of Wikipedia articles remains incredibly challenging~\cite{j.2018generating}. Particularly, articles categorized as B and C\footnote{\url{https://en.wikipedia.org/wiki/Wikipedia:Content_assessment}}, especially those on lesser-known biographies, often lack depth and detail. Enhancing these ``tail'' articles is crucial for providing comprehensive and accurate information to users, thus fulfilling Wikipedia's mission of offering reliable and detailed knowledge across all subjects.

\begin{figure}[!t]
    \centering
    \includegraphics[width=0.95\linewidth]{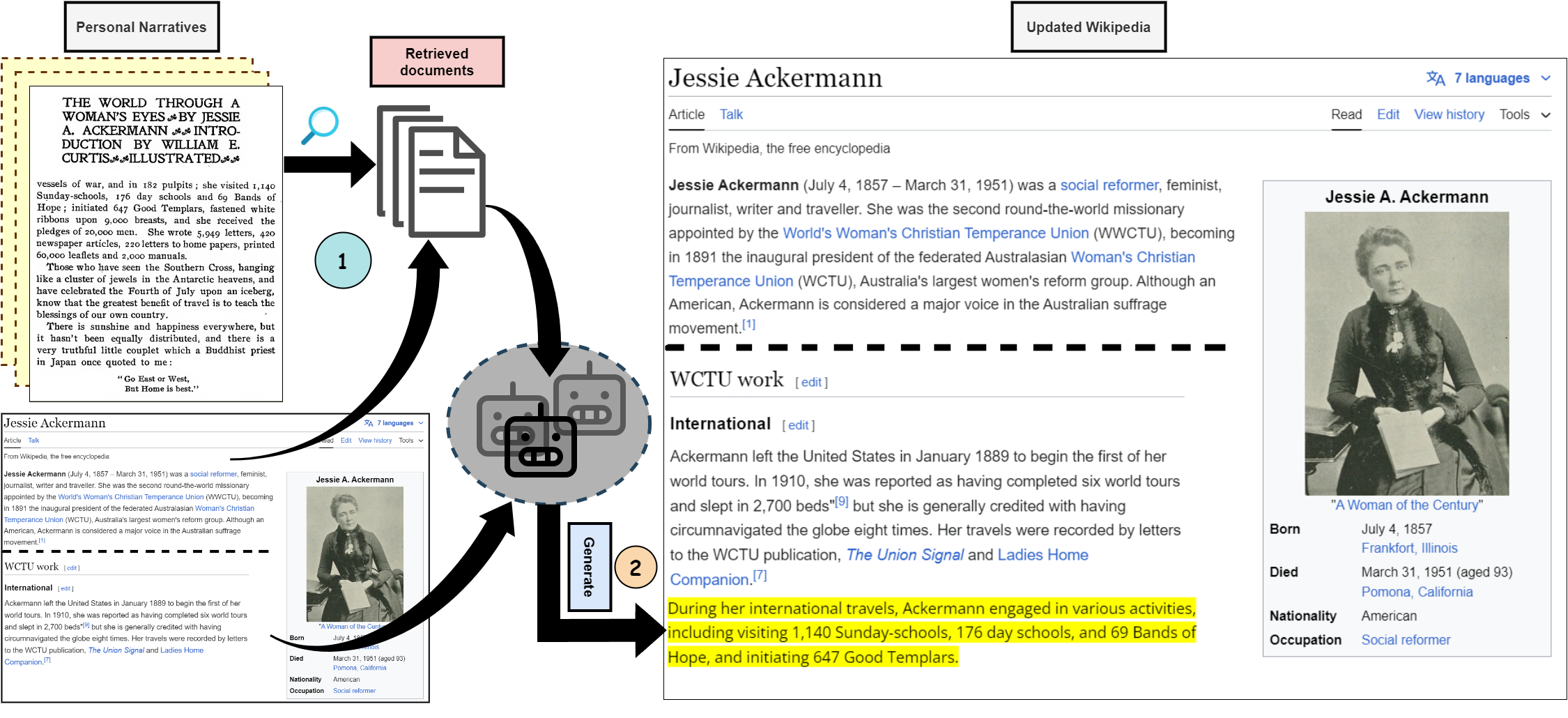}
    \caption{\footnotesize Overview of Wikipedia section enhancement from personal narratives.}
    \label{fig:intro}
\end{figure}
 Previous work on generating Wikipedia articles has generally focused on generating full Wikipedia article. For example, \citet{j.2018generating} assume that reference documents are provided in advance, while \citet{fan2022generating} assume an article outline is already available for generating full Wikpedia page. These assumptions do not hold universally, as the process of collecting references is inherently complex and resource-intensive. Moreover, these systems are not useful for updating existing texts as they can only generate text from scratch. \citet{iv-etal-2022-fruit} address this gap by proposing an approach to generate grounded text from given structured evidence to update existing text. This poses unique challenge as, the generated text needs to be faithful to both the original article and the external evidence, and determine which is relevant and which can be ignored.\\
To the best of our knowledge, none of the previous works specifically explore the use of personal narratives to enrich Wikipedia content.
Personal narratives offer a wealth of detailed, first-hand information. Autobiographies, as personal narratives, provide unique insights into individuals' consciousness and motivations, capturing historical details within the context of personal experiences~\cite{pascal2015design, popkin2005history, aurell2006autobiography}. Similarly, biographies, inherently tied to history, make the past more accessible and connected~\cite{caine2018biography, nature_of_biography}. By integrating rich, first-hand information from personal narratives, we aim to provide more comprehensive and accurate content. We presents a scalable solution for improving Wikipedia content quality, directly benefiting industries that rely on accurate knowledge bases, such as education, media, and digital libraries.
Our contributions are as follows:
\begin{compactitem}
    \item We propose a novel multi-staged approach \rvs{} to incorporate personal narratives, such as autobiographies and biographies to enhance Wikipedia tail articles, a problem which has not been extensively explored in previous research. 
    \item \textcolor{black}{We collect a large number of personal narratives relevant to the corresponding Wikipedia biography pages (53 for Class B, and 49 for Class C), which can be a good source of factually correct information.} 
    \item We rigorously evaluate the generated content using both automatic and crowd-based evaluations. Our method surpasses the standard RAG approach in readability, understandability, and information quality. Based on crowdsourced evaluation we find that \rvs{} substantially outperforms the best-performing baseline in terms of informativeness and integrability. Specifically, human judges mark 92\% of the generated content as integrable and 96\% as informative.
\end{compactitem}

\section{Related work}
\label{sec:related_work}
\noindent \textbf{Automatic Wikipedia article enhancement}: Automatic Wikipedia enhancement has been studied for more than a decade~\cite{10.1145/2682571.2797073,j.2018generating, fan2022generating, banerjee2016wikiwrite, zhang2024retrievalbasedfulllengthwikipediageneration}. In recent times, \citet{zhang2024retrievalbasedfulllengthwikipediageneration} leveraged RAG to create full length Wikipedia articles.\\ 
\noindent\textbf{Grounded content generation using RAG}: Augmenting language models (LMs) with retrieval at inference time is a typical way to leverage external knowledge stores~\citep{ram2023ralm,JMLR:v24:23-0037}. While some works use retrieval to construct demonstrations for in-context learning~\citep{poesia2022synchromesh,khattab2022demonstrate}, others~\cite{lewis2020retrieval,menick2022teaching,gao-etal-2023-enabling,bohnet2023attributed,qian2023webbrainlearninggeneratefactually} use retrieval to provide additional information for LMs to ground on.
While RAG is widely studied in question answering, how to use it for expanding a Wikipedia section is less investigated.\\
\noindent \textbf{Present work}: Although, there are several lines of work which are related to ours, none of them leverage personal narratives to improve Wikipedia articles. We carefully curate a set of autobiographies/biographies and develop algorithms so that the generated content is grounded on these narratives. In specific, we use a two-stage RAG pipeline for enhancing Wikipedia tail articles and outperform the most competing baseline.



\section{Data collection}
We employ a systematic approach to leverage autobiographical and biographical writings to enhance corresponding Wikipedia biography pages. This section details the process of selecting biographies and scraping biographical writings from digital libraries.\\ 
\noindent\textbf{Selecting biographies}: Wikipedia classifies its articles into several quality categories, such as FA (Featured Articles) and GA (Good Articles), A, B etc. For this study, we focus on biographies categorized as B and C. These categories represent articles that are informative but have significant scope for improvement. Our goal is to enrich these articles by integrating more comprehensive information. To begin with, we compile a list of titles from all B and C category biography articles on Wikipedia. This list serves as the basis for our subsequent scraping efforts. By targeting these specific categories, we aim to improve the quality and completeness of articles that currently lack sufficient information.\\   
\noindent\textbf{Scraping biographical writings}. We utilize online digital libraries, particularly \textit{Internet Archive}\footnote{\url{www.archive.org}}, to source the biographical writings required for our enhancements. \textit{Internet Archive} provides a vast collection of scanned historical books, making it an ideal resource for our purposes.\\
\noindent \textit{Automated search}: \textcolor{black}{To locate relevant biographical writings, we use the Internet Archive API\footnote{\url{https://archive.org/developers/quick-start-pip.html}}. For each name in our list of B and C class Wikipedia biographies, we search for the person name in the whole Internet Archive to retrieve the web link of the first item where textual content is available. These initial results are then subjected to a manual verification to filter out irrelevant and noisy links.} \\
\noindent \textit{Manual verification}: Due to the ambiguity in names and the nature of automated searches, many search results contain irrelevant or noisy information. To address this issue, we employ a post-graduate student who is a frequent Wikipedia user to manually verify the collected links. This step is crucial to ensure the quality and relevance of the biographical writings that we ultimately use. The manual verification process involves filtering noisy links that are not relevant to the specific Wikipedia biography or contain irrelevant information. We, then utilize the verified biographical writings to enrich the Wikipedia biography pages. By integrating detailed and reliable information from these sources, we aim to significantly improve the quality of the biographies on Wikipedia using the methodology described in Section~\ref{sec:method}.

\noindent\textbf{Dataset details}: Our dataset contains a total of 102 personal narratives (53 for Wikipedia class B, 49 for Wikipedia class C) from a diverse set of profiles. The detailed description of the personal narratives are noted in Table~\ref{tab:data_description}.



\label{sec:schema}
\begin{figure*}[!t]
    \centering
\includegraphics[width=0.95\linewidth]{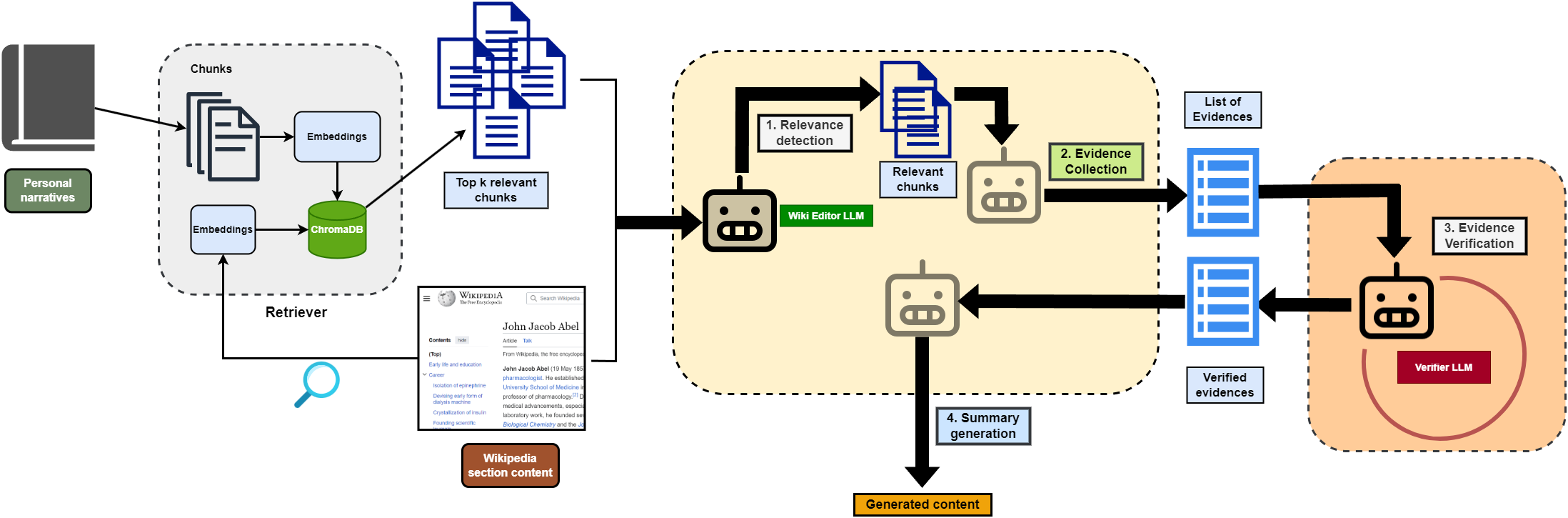}
    \caption{A schematic of \rvs{}. LLMs in the same block represents that they are in same chat session.}
    \label{fig:reversum}
\end{figure*}

\section{Task description}
\label{sec:task_def}

Our primary goal is to enhance biographical Wikipedia articles, especially those that are less comprehensive (B and C category articles), by leveraging personal narratives such as autobiographies and biographies. Consider, for a particular person $P$, $W_{P}$ is the Wikipedia page for $P$ consisting of $n$ sections, $W_{S_i}$ is the current section content for the section $S_i$, such that $W_P = \bigcup W_{S_{i}}$ where $i \in \{1..n\}$. Now, our goal is to utilize the personal narrative $B$ (e.g., biography) of $P$ to generate a text $G_{S_{i}}$ that is coherent with and relevant to $W_{S_{i}}$ such that the new content becomes $W'_{S_{i}}$, where $W'_{S_{i}} = W_{S_{i}} + G_{S_{i}}$.


\section{Methodology}
\label{sec:method}


\subsection{Pilot study with standard RAG}

We employ a standard RAG approach to enhance specific sections of biographical Wikipedia pages using corresponding personal narratives, such as autobiographies or biographies.\\
\noindent\textbf{Retriever}: Given a biographical Wikipedia page, we first consider the corresponding personal narrative (autobiography or biography) as the source of external knowledge. We, then split the text (i.e., personal narrative) into several chunks of fixed length (we vary the length $\in \{600, 800, 1000, 1200\}$ characters) with a window of 200 using \textit{RecusiveTextSplitter}\footnote{\url{https://python.langchain.com/v0.1/docs/modules/data\_connection/document\_transformers/recursive\_text\_splitter/}}. Following this we embedd each of the chunks using \texttt{sentence-bert}\footnote{\url{https://huggingface.co/sentence-transformers/all-mpnet-base-v2}} embeddings and store them in a vector database (we choose \textit{ChromaDB}\footnote{We also use other open-source vector stores - \textit{FAISS}, \textit{Pinecone} but do not observe significant difference.}). Subsequently, we curate a query consisting of the \textit{section title} and \textit{section content} of the Wikipedia article, and use maximum marginal relevance (MMR) based search to retrieve top $k$ chunks (we vary $k \in \{2-5\}$) relevant to the query. \\
\noindent\textbf{Generation and section enhancement}: We use several state-of-the-art large language models (LLM) to perform text generation. This generated text can be appended to an existing Wikipedia section. First, we carefully design a prompt which consist of two inputs - (1) the existing content of the Wikipedia section, (2) retrieved context (top k chunks relevant to retrieval query) and an instruction. The exact prompt can be found in Table~\ref{app:prompt_std_RAG} of Appendix~\ref{sec:prompts}.\\
\noindent\textbf{Generated content analysis}: As, LLM generated contents are oftentimes prone to hallucination, there is a need for manual verification for the content. We randomly select 100 Wikipedia sections and the corresponding generated content to evaluate the quality of the content. The evaluation was done by 9 Wikipedia users including an expert in Wikipedia research all of whom voluntarily participated in the task. We ask the participants whether the generated content can be integrated with the existing content or not, and a free text field to fill any concern about the generated content. We observe that, overall, in 56\% cases the participants mentioned that the generated contents are just a summary of the already existing Wikipedia content. This demonstrates that a simple RAG based generation pipeline might not be an accurate choice for this task.

\subsection{\rvs{}}
In this setup we propose a multi-staged generation approach containing \underline{R}elevance detection, \underline{E}vidence collection, \underline{Ver}ification, and \underline{Sum}marization -- \rvs{}, which aims to reduce redundant information and ensure the generation of grounded and accurate content from personal narratives. A schematic of \rvs{} is presented in Figure~\ref{fig:reversum}.\\
In the retrieval phase we use the same technique as the initial RAG based approach. Before the generation we execute the following steps.\\
\noindent \textbf{Relevance detection}: The first stage of \rvs{} comprises an LLM, used for identifying the most relevant chunk out of the top $k$ retrieved chunks from the retrieval phase for a specific section content. We use the specific section content and the retrieved chunks as input to the LLM, and ask to respond only the most relevant chunks based on the section content. We provide the privilege to the LLM to produce `No relevant chunks' in case it thinks there is no chunk related to the section content. The exact prompt for this relevance detection phase is shown in Table~\ref{app:prompt_relevant_detection} of Appendix~\ref{sec:prompts}.\\
\noindent \textbf{Evidence collection}:  In this second step, we select evidences from the most relevant documents identified in the previous step. We use the previous chat history, while performing the evidence collection step. This step yields a list of evidences (specific phrases) from the retrieved chunks. The exact prompt for selecting the evidence can be found in Table~\ref{app:prompt_evidence_extraction} of Appendix~\ref{sec:prompts}.\\ 
\noindent \textbf{Verification}: \textcolor{black}{The verification stage ensures that the extracted evidences originate solely from the retrieved chunks, maintaining the integrity and reliability of the information. To mitigate hallucinations, we use a \underline{separate chat session} for this phase. During verification, the input to the LLM contains only the ``retrieved chunks'' and ``extracted evidences'' from the source material, with no extraneous information. The LLM verifies whether each evidence is present in the retrieved chunks, ensuring no external or unsupported information is introduced. This process results in a list of evidences confirmed to be from the retrieved chunks, guaranteeing their relevance and accuracy. The prompt for verification can be found in Table~\ref{app:prompt_evidence_verification} of Appendix~\ref{sec:prompts}}.\\
\noindent \textbf{Summarization}: In the final stage, the LLM generates a summarized content from the verified evidences, ensuring seamless integration with the existing section content. We provide the LLM with the verified evidences and instruct it to generate a concise and coherent summary based on these evidences. The summary is designed to integrate seamlessly with the existing content of the Wikipedia section. The prompt for verification can be found in Table~\ref{app:prompt_summary_generation} of Appendix~\ref{sec:prompts}. We use \texttt{Llama-3-8b-instruct} model as the LLM. The implementation details and hyperparameters can be found in Appendix~\ref{appendix:model_implementation}.

\subsection{Handling negative scenario}
In some cases, it is possible that from the retrieved context the particular Wikipedia section cannot be expanded due to semantic or factual differences. We handle such cases, using two approaches.
\noindent\textbf{Thresholding in retrieval}: The retrieved contexts are generally based on the semantic similarity between the existing Wikipedia section content and the chunks from the personal narratives. We apply a threshold similarity value of 0.3\footnote{We apply a grid search of sets of 0.1 to select this particular value.}, only beyond which we consider expanding the particular section from the retrieved contexts.  \\
\noindent\textbf{Using prompting}: Sometimes, top semantically similar retrieved contexts may not be appropriate for expanding particular Wikipedia section. To tackle such scenarios, we use an appropriate prompt which can tell whether the Wikipedia section can be expanded or not from the retrieved contexts during the generation phase.

\subsection{Baselines}
\textcolor{black}{There is no recent work that directly addresses the specific task of enhancing lesser-known Wikipedia biographies. Most contemporary approaches focus either on generating full-length Wikipedia articles using web-based sources \cite{zhang2024retrievalbasedfulllengthwikipediageneration, shao2024assisting}, or augmenting content related to well known events \cite{iv-etal-2022-fruit}. \citet{banerjee2015wikikreator} worked on enhancing Wikipedia stubs. To provide a broader baseline, we implemented an approach inspired by \citet{banerjee2015wikikreator}, tailored to our use case. Rather than web-based retrieval, we employ a vector store retrieval to obtain similar documents and integrate a more advanced summarization technique using a generative model (\textsc{LLaMA-3}). In contrast, \citet{banerjee2015wikikreator} used integer linear programming (ILP)-based abstractive summarization. In addition, we propose two strong baselines along with \rvs{}.}\\
\noindent\textbf{Key-phrase extraction from personal narrative}: We split the personal narratives into chapters and extract key phrases using three techniques: (i) KeyBert~\cite{grootendorst2020keybert}, (ii) Yake~\cite{campos2020yake} and, (iii) Rakun2~\cite{vskrlj2022retrieval}. From each chapter, we extract five key phrases, varying the number of words (1-3). 

\noindent\textbf{Key-phrase focused paragraphs}: We generate paragraphs relevant to each key phrase using two methods:\\
\noindent \textit{1. Coherence score \cite{jwalapuram-etal-2022-rethinking} based}: Sentences from the chapters are split using sentence breaks and encoded with \texttt{sentence-bert} embeddings. We select the top 20 sentences based on cosine similarity to the key phrase. A paragraph is initialized with the most similar sentence, and sentences are appended if the coherence score improves.\\
\noindent \textit{2. RAG-based}: We use key phrases as queries to retrieve top chunks from the narratives. An LLM then generates a paragraph from these chunks.\\
\noindent\textbf{Wiki-section to key-phrases map}: We map the key phrases ($kp$) and their focused paragraphs ($P$) to Wikipedia sections ($S$). Using \texttt{sentence-bert}, we encode key phrases, paragraphs, and sections, measuring similarity through three features: cosine similarity between section and key-phrase embeddings, section and paragraph embeddings, and key-phrase and paragraph embeddings. The final similarity between a section $S_i$ and a key-phrase $kp_j$is given by: $\alpha * sim(S_i, kp_j) - \beta * sim(S_i, P_j) + \gamma * sim(kp_j, P_j)$  
 where $\alpha$, $\beta$, and $\gamma$ are hyperparameters. The expression attempts to select those paragraphs ($P_j$) that are similar to the key-phrases but at the same time distant from the section content to avoid inclusion of redundant information. More experimental details about the baselines are provided in Appendix~\ref{appendix:baseline_details}.

\subsection{Evaluation metrics} 
Most of the previous evaluation strategies such as ORES\footnote{\url{https://www.mediawiki.org/wiki/ORES}} employ Wikipedia revision ids for evaluating the quality of a Wikipedia page. However, in our case this approach is not applicable. A more suitable metric has been suggested in~\cite{9770051}, which includes $E$ (Expertise), $A$ (Authority), and $T$ (Trustworthiness). However, we had to exclude $A$ and $T$ as these are dependent on page links, number of edits, since we are only adding the textual content. $E$ is measured in terms of the Quality of a Wikipedia page content which is defined as: 
\footnotesize
\underline{Quality} = 0.255 * Informativeness + 0.654 * Readability +  0.557 * Understandability. Informativeness represents the size of the textual content present in the Wikipedia page, readability and understandability provide insights about the linguistic quality and are defined as:\\
\underline{Informativeness} = 0.12 * \texttt{page-size} + 0.151 * \texttt{\#sentences} + 0.154 * \texttt{\#words} + 0.155 * \texttt{\#complex-words};\\
\noindent \underline{Readability} = 0.213 * \texttt{Flesch-Kincaid-grade-evel}
+ 0.185 * \texttt{Coleman-Liau-index} + 0.26 * \texttt{\%complex-words} + 0.253 * \texttt{avg-syllables-per-word};\\ 
\noindent \underline{Understandability} = 0.393 * \texttt{Gunning-Fog-score} + 0.352 * \texttt{SMOG-index} + 0.181 * \texttt{automated-readability-index} + 0.344 * \texttt{avg-words-per-sentence}; \\
\normalsize
\noindent We measure the relative improvement as: $\varDelta Quality = Quality(W_{S} + G_{S}) - Quality(W_{S})$. However, the simple `Informativeness' metric does not take into the account (a) how much new information has been added, and (b) how much appropriate the content is in continuing the existing section. To tackle this, we propose a `Calibrated Informativeness (CI)',  formally defined as: $\varDelta CI = \varDelta Informativeness * \texttt{fraction-of-newly-added-words} \quad * \quad \texttt{continuation-score}$ where, the fraction of new added words determines how much new information has been added, and the continuation score determines how much the new content is appropriate in expanding the existing section content. To measure the continuation score we employ a supervised approach by fine-tuning a \texttt{Llama-3-8b-instruct} model. The fine-tuning strategy is discussed in details in Appendix~\ref{appendix:calibrated_informativeness}.
\begin{table}[H]\centering
\renewcommand{\arraystretch}{1.2}
\resizebox{\columnwidth}{!}{
\begin{tabular}{l|l|c|c|c|c}\toprule
\textbf{Wikipedia class} &\textbf{Method} &\textbf{$\varDelta CI$} &\textbf{$\varDelta Und.$} &\textbf{$\varDelta Read.$} &\textbf{$\varDelta Quality$} \\\midrule
\multirow{5}{*}{class B}
&\citet{banerjee2015wikikreator}* & 23.23	&-0.35	&-0.03	&5.71 \\
&\textit{Key-phrase to section mapping (Coherence score based)} &57.26 &-0.62 &0.01 &14.2 \\
&\textit{Key-phrase to section mapping (RAG based)} &51.5 &-0.28 &0.03 &12.94 \\
&Standard RAG &49.29 &-0.08 &-0.01 &12.51 \\
&\textbf{\rvs{}} &\cellcolor{ForestGreen}\textbf{61.27} &\cellcolor{ForestGreen}\textbf{0.27} &\cellcolor{ForestGreen}\textbf{0.10} &\cellcolor{ForestGreen}\textbf{15.84} \\\hline
\multirow{5}{*}{class C}
&\citet{banerjee2015wikikreator}* & 18.8	&0.24	&-0.01	&4.94 \\
&\textit{Key-phrase to section mapping (Coherence score based)} &8.34 &-0.23 &0.04 &2.0 \\
&\textit{Key-phrase to section mapping (RAG based)} &7.38 &-0.11 &0.03 &1.83 \\
&Standard RAG &38.61 &0.29 &\cellcolor{ForestGreen}\textbf{0.14} &10.12 \\
&\textbf{\rvs{}} & \cellcolor{ForestGreen}\textbf{59.26} &\cellcolor{ForestGreen}\textbf{0.35} &0.08 &\cellcolor{ForestGreen}\textbf{13.00} \\
\bottomrule
\end{tabular}
}
\caption{\footnotesize Comparative results for \rvs{} with other baselines. The metrics are averaged across all biographies for each Wikipedia class. The best results are in \textbf{boldface} and \colorbox{ForestGreen}{highlighted}. * We use a modified implementation of \citet{banerjee2015wikikreator}. \label{tab:comparative_result}}
\end{table}

\section{Results}

The key results are subdivided based on two ways of evaluation -- automatic and manual.\\
\noindent\textbf{Automatic evaluation}: \textcolor{black}{We report the results of the automatic evaluation in Table~\ref{tab:comparative_result}. In terms of average overall quality as well as in terms of all the individual component averages, \rvs{} substantially outperforms the other baselines for the class B articles. For the class C articles, while the average overall quality is again best for \rvs{}, it only slightly underperforms in terms of average readability. We conduct a Mann-Whitney U-test to compare the \rvs{}-based results with the best-performing baseline (standard RAG-based) for both B and C category articles. For the B category, we observe statistically significant improvements ($p$-value $< 0.05$) across all four metrics: understandability, readability, calibrated informativeness, and quality. For the C category, statistically significant improvements ($p$-value $< 0.05$) were observed for calibrated informativeness and quality. The results for each individual article is noted in Table~\ref{tab:individual_result_llama-3} of Appendix~\ref{appendix:results}.}\\

\noindent\textbf{Manual evaluation}: We randomly select 100 Wikipedia section and the corresponding generated content from \rvs{} for the manual evaluation\footnote{We compensate the annotators with a \$4 amazon gift voucher each.}. We employ 8 individuals from a diverse backgrounds to manually verify the generated content. 
For each of the samples (existing Wikipedia section and the generated content), we first ask whether the generated content can be seamlessly integrated with the existing Wikipedia section followed by a few questions related to informativeness, understandability, and readability. We obtain two judgments per sample. We observe that in a total of 92\% cases the annotators marked `yes' for whether the generated content can be integrated with the existing section (Cohen's $\kappa$ score of 0.84). Similarly, in 96\%, 98\%, and 99\% cases the annotators found the generated contents are informative, understandable, and readable respectively. Also there was no case where the annotator raised concern about generating duplicate information from the existing section. For the best performing baseline in terms of automatic evaluation (i.e., standard RAG based approach) the number of cases where the annotators marked yes is 75\% for the integrability, while in 67.5\%, 98\%, and 98\% cases the annotators found the generated contents are informative, understandable, and readable respectively. \textcolor{black}{In addition, we obtain a GPT-4 based \textit{faithfulness \cite{es-etal-2024-ragas} score} of 0.95 for the \rvs{} generated summary with respect to the content from the personal narratives. The details of the evaluation of the generated summary are provided in Appendix~\ref{appendix:factual_correctness}.} These results together portray the overall impressive performance of the \rvs{}\footnote{Some failure cases are discussed in Appendix~\ref{analysis:negative_scenario}.}.  
\section{Analysis}
\label{sec:analysis}
\subsection{Analysis of the negative scenario}
\label{analysis:negative_scenario}
\textcolor{black}{We analyze the cases where the overall pipeline is not able to generate a coherent content that can be integrated with the existing Wikipedia section. This can happen due to the poor semantic relation of the retrieved chunks from the personal narratives with the section content or the \rvs{} pipeline finding insufficient information to enhance the existing content. We observe that, in around 16\% cases the retrieved documents are less similar (than threshold value of 0.3) to the existing content. In around 35\% cases, the \rvs{} pipeline judges the retrieved information is not sufficient to expand the existing section content. Each stage-wise details are provided in Table~\ref{tab:non_expansible_case_numeric}. Further analysis is presented in Appendix~\ref{sec:additional_analysis}.}
\begin{table}[H]
\centering
\scriptsize
\begin{tabular}{c|c}\hline
     \textbf{Reason for non-expansion} &\textbf{Percentage}  \\\hline
     Retrieval & 16\%\\
     Relevance detection & 12\%\\
     Evidence collection & 3\%\\
     Evidence verification & 19\%\\
     Summary generation & 1\% \\
     \hline
     
\end{tabular}
\caption{\footnotesize Stage-wise percentages of non-expansible cases.}
\label{tab:non_expansible_case_numeric}
\end{table}

\subsection{Which portion of the narrative is important for which section}
\begin{figure}[H]
    \centering
\includegraphics[width=\columnwidth]{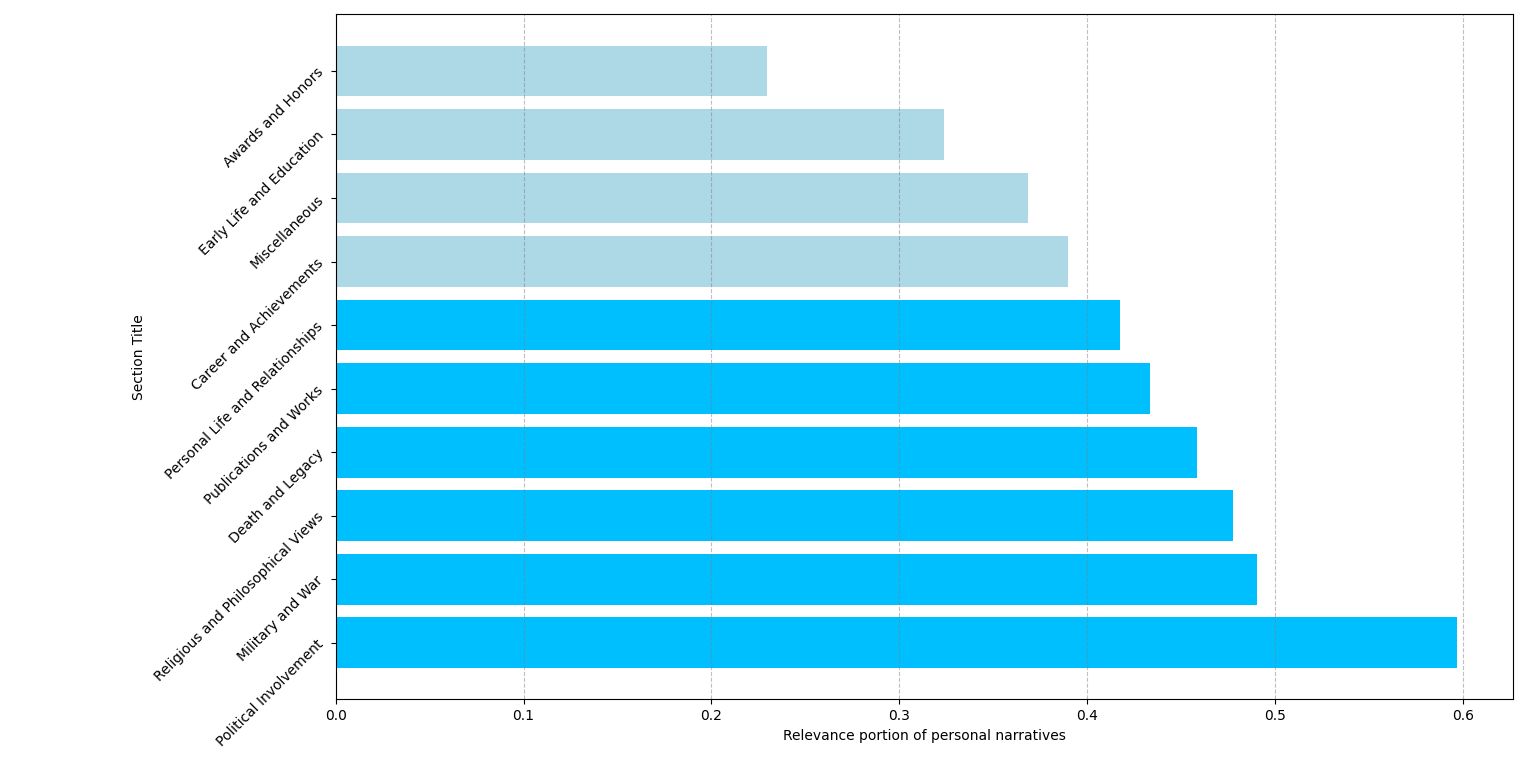}
    \caption{\footnotesize Relevance of different portions of the personal narratives with respect to the Wikipedia section.}
    \label{fig:relevant_portion}
\end{figure}

\textcolor{black}{We aim to observe which part of the input personal narratives are
more crucial in expanding which Wikipedia section. During the retrieval of context we utilize the relative position of the divided chunks to understand the positional relevance of the particular chunk in the personal narrative with respect to the particular Wikipedia section. We divide all the section titles to 10 predefined categories and plot the average relative position of the retrieved chunks. The plot is shown in Figure~\ref{fig:relevant_portion}. We notice that, the initial portions of the personal narratives are relevant to the sections such as `Early life', `Education', and `Awards and Honors', whereas the later portion of the personal narratives are more related to the sections like `Political involvement' and `Military activities'.}

\section{Ablation study}
\label{sec:ablation_study}
\textcolor{black}{We use ablation to understand the effectiveness of each stage in \rvs{}. We show the average results (for both B ad C class books) in Table~\ref{tab:ablation}. We can observe that, without \textit{Evidence verification} stage, the quality of the generated content reduce drastically.}

\begin{table}[H]\centering
\resizebox{\columnwidth}{!}{
\begin{tabular}{l|c|cccc}\toprule
& &\textbf{$\varDelta CI$} &\textbf{$\varDelta Understandability$} &\textbf{$\varDelta Readability$} &\textbf{$\varDelta Quality$} \\\midrule
&Actual &60.27 &0.31 &0.09 &14.41 \\
&w/o Relevance detection &55.40 &0.36 &0.04 &14.38  \\
\rvs{} &w/o Evidence collection &51.33 &0.17 &0.03 &13.22 \\
&w/o Evidence verification &47.25 &0.23 &0.03 & 12.22\\
&w/o Summary generation &52.89 &0.07 &0.02 & 13.54\\
\bottomrule
\end{tabular}
}
\caption{\footnotesize Results without different stages in \rvs{}. Note that in \textit{w/o Evidence collection} stage we did not consider the verification.}\label{tab:ablation}
\end{table}

\section{Additional details}
\paragraph{Generalizing \rvs{} for other Wikipedia article types.} \textcolor{black}{Our approach is specifically tailored for Wikipedia tail articles, focusing on sequentially enhancing their sections. Currently, we limit our methodology to B and C classes, as lower-category articles often lack well-defined sections. In future, we aim to explore how this approach can be generalized to accommodate a broader range of Wikipedia article types.}

\paragraph{Inter-section redundancy of the generated content.} \textcolor{black}{Our current methodology independently enhances each Wikipedia section, and we do not explicitly measure inter-section alignment or ensure consistency across sections. To avoid duplication across other sections, our system relies on section-specific relevance cues during retrieval and evidence selection. However, we acknowledge that ensuring absolute non-duplication across all sections is challenging. Future work could explore inter-section alignment strategies to refine this process further and ensure maximal informativeness while minimizing overlap.}

\section{Conclusion}
In this study, we introduced \rvs{}, a novel multi-staged RAG pipeline to enhance Wikipedia biographies of lesser-known individuals using personal narratives. Our approach systematically incorporates relevance detection, evidence collection, verification, and summarization to ensure the generation of accurate and informative content. Through rigorous evaluation, both automatic and manual, we demonstrated that \rvs{} substantially outperforms the traditional RAG-based methods. 
\section{Limitations}
Despite the promising results, our study has certain limitations. First, the reliance on personal narratives such as autobiographies/biographies may introduce a subjective bias, as these sources often reflect personal perspectives and interpretations which could be in conflict with Wikipedia's neutral point of view policy. In addition, our manual verification process, while necessary to ensure content quality, is inherently subjective and may lead to inconsistencies in the evaluation of relevance and accuracy. The dataset of personal narratives, though diverse, may not be representative of all lesser-known biographies, potentially limiting the generalizability of our approach. Future research should explore the integration of more diverse sources and the development of automated verification techniques to address these limitations.
\section{Ethical considerations}
The biographical writings used for data collection were sourced from publicly available digital libraries, ensuring compliance with copyright policies and respect for intellectual property rights. We ensured that all human annotators involved in the manual verification process participated voluntarily and provided informed consent. No personally identifiable information was collected from the annotators, preserving their anonymity and privacy. Further, we took every measure to avoid the inclusion of any sensitive or potentially harmful content in the enhanced Wikipedia articles. 

\bibliography{custom}

\begin{thebibliography}{35}
\providecommand{\natexlab}[1]{#1}

\bibitem[{Aurell(2006)}]{aurell2006autobiography}
Jaume Aurell. 2006.
\newblock Autobiography as unconventional history: Constructing the author.
\newblock \emph{Rethinking History}, 10(3):433--449.

\bibitem[{Banerjee and Mitra(2015{\natexlab{a}})}]{10.1145/2682571.2797073}
Siddhartha Banerjee and Prasenjit Mitra. 2015{\natexlab{a}}.
\newblock \href {https://doi.org/10.1145/2682571.2797073} {Filling the gaps: Improving wikipedia stubs}.
\newblock In \emph{Proceedings of the 2015 ACM Symposium on Document Engineering}, DocEng '15, page 117–120, New York, NY, USA. Association for Computing Machinery.

\bibitem[{Banerjee and Mitra(2015{\natexlab{b}})}]{banerjee2015wikikreator}
Siddhartha Banerjee and Prasenjit Mitra. 2015{\natexlab{b}}.
\newblock Wikikreator: Improving wikipedia stubs automatically.
\newblock In \emph{Proceedings of the 53rd Annual Meeting of the Association for Computational Linguistics and the 7th International Joint Conference on Natural Language Processing (Volume 1: Long Papers)}, pages 867--877.

\bibitem[{Banerjee and Mitra(2016)}]{banerjee2016wikiwrite}
Siddhartha Banerjee and Prasenjit Mitra. 2016.
\newblock Wikiwrite: Generating wikipedia articles automatically.
\newblock In \emph{IJCAI}, pages 2740--2746.

\bibitem[{Bohnet et~al.(2023)Bohnet, Tran, Verga, Aharoni, Andor, Soares, Ciaramita, Eisenstein, Ganchev, Herzig, Hui, Kwiatkowski, Ma, Ni, Saralegui, Schuster, Cohen, Collins, Das, Metzler, Petrov, and Webster}]{bohnet2023attributed}
Bernd Bohnet, Vinh~Q. Tran, Pat Verga, Roee Aharoni, Daniel Andor, Livio~Baldini Soares, Massimiliano Ciaramita, Jacob Eisenstein, Kuzman Ganchev, Jonathan Herzig, Kai Hui, Tom Kwiatkowski, Ji~Ma, Jianmo Ni, Lierni~Sestorain Saralegui, Tal Schuster, William~W. Cohen, Michael Collins, Dipanjan Das, Donald Metzler, Slav Petrov, and Kellie Webster. 2023.
\newblock \href {https://arxiv.org/abs/2212.08037} {Attributed question answering: Evaluation and modeling for attributed large language models}.
\newblock \emph{Preprint}, arXiv:2212.08037.

\bibitem[{Brown et~al.(2020)Brown, Mann, Ryder, Subbiah, Kaplan, Dhariwal, Neelakantan, Shyam, Sastry, Askell et~al.}]{brown2020language}
Tom Brown, Benjamin Mann, Nick Ryder, Melanie Subbiah, Jared~D Kaplan, Prafulla Dhariwal, Arvind Neelakantan, Pranav Shyam, Girish Sastry, Amanda Askell, et~al. 2020.
\newblock Language models are few-shot learners.
\newblock \emph{Advances in neural information processing systems}, 33:1877--1901.

\bibitem[{Caine(2018)}]{caine2018biography}
Barbara Caine. 2018.
\newblock \emph{Biography and history}.
\newblock Macmillan international higher education.

\bibitem[{Campos et~al.(2020)Campos, Mangaravite, Pasquali, Jorge, Nunes, and Jatowt}]{campos2020yake}
Ricardo Campos, V{\'\i}tor Mangaravite, Arian Pasquali, Al{\'\i}pio Jorge, C{\'e}lia Nunes, and Adam Jatowt. 2020.
\newblock Yake! keyword extraction from single documents using multiple local features.
\newblock \emph{Information Sciences}, 509:257--289.

\bibitem[{Chen et~al.(2017)Chen, Fisch, Weston, and Bordes}]{chen-etal-2017-reading}
Danqi Chen, Adam Fisch, Jason Weston, and Antoine Bordes. 2017.
\newblock \href {https://doi.org/10.18653/v1/P17-1171} {Reading {W}ikipedia to answer open-domain questions}.
\newblock In \emph{Proceedings of the 55th Annual Meeting of the Association for Computational Linguistics (Volume 1: Long Papers)}, pages 1870--1879, Vancouver, Canada. Association for Computational Linguistics.

\bibitem[{Es et~al.(2024)Es, James, Espinosa~Anke, and Schockaert}]{es-etal-2024-ragas}
Shahul Es, Jithin James, Luis Espinosa~Anke, and Steven Schockaert. 2024.
\newblock \href {https://aclanthology.org/2024.eacl-demo.16} {{RAGA}s: Automated evaluation of retrieval augmented generation}.
\newblock In \emph{Proceedings of the 18th Conference of the European Chapter of the Association for Computational Linguistics: System Demonstrations}, pages 150--158, St. Julians, Malta. Association for Computational Linguistics.

\bibitem[{Fan and Gardent(2022)}]{fan2022generating}
Angela Fan and Claire Gardent. 2022.
\newblock Generating biographies on wikipedia: The impact of gender bias on the retrieval-based generation of women biographies.
\newblock In \emph{Proceedings of the 60th Annual Meeting of the Association for Computational Linguistics (Volume 1: Long Papers)}, pages 8561--8576.

\bibitem[{Gao et~al.(2023)Gao, Yen, Yu, and Chen}]{gao-etal-2023-enabling}
Tianyu Gao, Howard Yen, Jiatong Yu, and Danqi Chen. 2023.
\newblock \href {https://aclanthology.org/2023.emnlp-main.398} {Enabling large language models to generate text with citations}.
\newblock In \emph{Proceedings of the 2023 Conference on Empirical Methods in Natural Language Processing}, pages 6465--6488, Singapore. Association for Computational Linguistics.

\bibitem[{Garraty(1957)}]{nature_of_biography}
John~A. Garraty. 1957.
\newblock \href {http://www.jstor.org/stable/45133792} {The nature of biography}.
\newblock \emph{The Centennial Review of Arts \& Science}, 1(2):123--141.

\bibitem[{Grootendorst(2020)}]{grootendorst2020keybert}
Maarten Grootendorst. 2020.
\newblock \href {https://doi.org/10.5281/zenodo.4461265} {Keybert: Minimal keyword extraction with bert.}

\bibitem[{Iv et~al.(2022)Iv, Passos, Singh, and Chang}]{iv-etal-2022-fruit}
Robert Iv, Alexandre Passos, Sameer Singh, and Ming-Wei Chang. 2022.
\newblock \href {https://doi.org/10.18653/v1/2022.naacl-main.269} {{FRUIT}: Faithfully reflecting updated information in text}.
\newblock In \emph{Proceedings of the 2022 Conference of the North American Chapter of the Association for Computational Linguistics: Human Language Technologies}, pages 3670--3686, Seattle, United States. Association for Computational Linguistics.

\bibitem[{Izacard et~al.(2023)Izacard, Lewis, Lomeli, Hosseini, Petroni, Schick, Dwivedi-Yu, Joulin, Riedel, and Grave}]{JMLR:v24:23-0037}
Gautier Izacard, Patrick Lewis, Maria Lomeli, Lucas Hosseini, Fabio Petroni, Timo Schick, Jane Dwivedi-Yu, Armand Joulin, Sebastian Riedel, and Edouard Grave. 2023.
\newblock \href {http://jmlr.org/papers/v24/23-0037.html} {Atlas: Few-shot learning with retrieval augmented language models}.
\newblock \emph{Journal of Machine Learning Research}, 24(251):1--43.

\bibitem[{Jwalapuram et~al.(2022)Jwalapuram, Joty, and Lin}]{jwalapuram-etal-2022-rethinking}
Prathyusha Jwalapuram, Shafiq Joty, and Xiang Lin. 2022.
\newblock \href {https://doi.org/10.18653/v1/2022.acl-long.418} {Rethinking self-supervision objectives for generalizable coherence modeling}.
\newblock In \emph{Proceedings of the 60th Annual Meeting of the Association for Computational Linguistics (Volume 1: Long Papers)}, pages 6044--6059, Dublin, Ireland. Association for Computational Linguistics.

\bibitem[{Khattab et~al.(2022)Khattab, Santhanam, Li, Hall, Liang, Potts, and Zaharia}]{khattab2022demonstrate}
Omar Khattab, Keshav Santhanam, Xiang~Lisa Li, David Hall, Percy Liang, Christopher Potts, and Matei Zaharia. 2022.
\newblock Demonstrate-search-predict: Composing retrieval and language models for knowledge-intensive {NLP}.
\newblock \emph{arXiv preprint arXiv:2212.14024}.

\bibitem[{Kirchenbauer and Barns()}]{kirchenbauer2024hallucination}
Jason Kirchenbauer and Caleb Barns.
\newblock Hallucination reduction in large language models with retrieval-augmented generation using wikipedia knowledge.

\bibitem[{Lewis et~al.(2020)Lewis, Perez, Piktus, Petroni, Karpukhin, Goyal, K{\"u}ttler, Lewis, Yih, Rockt{\"a}schel et~al.}]{lewis2020retrieval}
Patrick Lewis, Ethan Perez, Aleksandra Piktus, Fabio Petroni, Vladimir Karpukhin, Naman Goyal, Heinrich K{\"u}ttler, Mike Lewis, Wen-tau Yih, Tim Rockt{\"a}schel, et~al. 2020.
\newblock Retrieval-augmented generation for knowledge-intensive nlp tasks.
\newblock \emph{Advances in Neural Information Processing Systems}, 33:9459--9474.

\bibitem[{Liu et~al.(2018)Liu, Saleh, Pot, Goodrich, Sepassi, Kaiser, and Shazeer}]{j.2018generating}
Peter~J. Liu, Mohammad Saleh, Etienne Pot, Ben Goodrich, Ryan Sepassi, Lukasz Kaiser, and Noam Shazeer. 2018.
\newblock \href {https://openreview.net/forum?id=Hyg0vbWC-} {Generating wikipedia by summarizing long sequences}.
\newblock In \emph{International Conference on Learning Representations}.

\bibitem[{Menick et~al.(2022)Menick, Trebacz, Mikulik, Aslanides, Song, Chadwick, Glaese, Young, Campbell-Gillingham, Irving, and McAleese}]{menick2022teaching}
Jacob Menick, Maja Trebacz, Vladimir Mikulik, John Aslanides, Francis Song, Martin Chadwick, Mia Glaese, Susannah Young, Lucy Campbell-Gillingham, Geoffrey Irving, and Nat McAleese. 2022.
\newblock \href {https://arxiv.org/abs/2203.11147} {Teaching language models to support answers with verified quotes}.
\newblock \emph{Preprint}, arXiv:2203.11147.

\bibitem[{Nogueira et~al.(2020)Nogueira, Jiang, Pradeep, and Lin}]{nogueira-etal-2020-document}
Rodrigo Nogueira, Zhiying Jiang, Ronak Pradeep, and Jimmy Lin. 2020.
\newblock \href {https://doi.org/10.18653/v1/2020.findings-emnlp.63} {Document ranking with a pretrained sequence-to-sequence model}.
\newblock In \emph{Findings of the Association for Computational Linguistics: EMNLP 2020}, pages 708--718, Online. Association for Computational Linguistics.

\bibitem[{Pascal(2015)}]{pascal2015design}
Roy Pascal. 2015.
\newblock \emph{Design and truth in autobiography}.
\newblock Routledge.

\bibitem[{Poesia et~al.(2022)Poesia, Polozov, Le, Tiwari, Soares, Meek, and Gulwani}]{poesia2022synchromesh}
Gabriel Poesia, Alex Polozov, Vu~Le, Ashish Tiwari, Gustavo Soares, Christopher Meek, and Sumit Gulwani. 2022.
\newblock \href {https://openreview.net/forum?id=KmtVD97J43e} {Synchromesh: Reliable code generation from pre-trained language models}.
\newblock In \emph{International Conference on Learning Representations}.

\bibitem[{Popkin(2005)}]{popkin2005history}
Jeremy~D Popkin. 2005.
\newblock \emph{History, historians, and autobiography}.
\newblock University of Chicago Press.

\bibitem[{Qian et~al.(2023)Qian, Zhu, Dou, Gu, Zhang, Liu, Lai, Cao, Nie, and Wen}]{qian2023webbrainlearninggeneratefactually}
Hongjing Qian, Yutao Zhu, Zhicheng Dou, Haoqi Gu, Xinyu Zhang, Zheng Liu, Ruofei Lai, Zhao Cao, Jian-Yun Nie, and Ji-Rong Wen. 2023.
\newblock \href {https://arxiv.org/abs/2304.04358} {Webbrain: Learning to generate factually correct articles for queries by grounding on large web corpus}.
\newblock \emph{Preprint}, arXiv:2304.04358.

\bibitem[{Ram et~al.(2023)Ram, Levine, Dalmedigos, Muhlgay, Shashua, Leyton-Brown, and Shoham}]{ram2023ralm}
Ori Ram, Yoav Levine, Itay Dalmedigos, Dor Muhlgay, Amnon Shashua, Kevin Leyton-Brown, and Yoav Shoham. 2023.
\newblock \href {https://arxiv.org/abs/2302.00083} {In-context retrieval-augmented language models}.
\newblock \emph{Transactions of the Association for Computational Linguistics}.

\bibitem[{Reid et~al.(2022)Reid, Yamada, and Gu}]{reid2022can}
Machel Reid, Yutaro Yamada, and Shixiang~Shane Gu. 2022.
\newblock Can wikipedia help offline reinforcement learning?
\newblock \emph{arXiv preprint arXiv:2201.12122}.

\bibitem[{Shao et~al.(2024)Shao, Jiang, Kanell, Xu, Khattab, and Lam}]{shao2024assisting}
Yijia Shao, Yucheng Jiang, Theodore Kanell, Peter Xu, Omar Khattab, and Monica Lam. 2024.
\newblock Assisting in writing wikipedia-like articles from scratch with large language models.
\newblock In \emph{Proceedings of the 2024 Conference of the North American Chapter of the Association for Computational Linguistics: Human Language Technologies (Volume 1: Long Papers)}, pages 6252--6278.

\bibitem[{{\v{S}}krlj et~al.(2022){\v{S}}krlj, Koloski, and Pollak}]{vskrlj2022retrieval}
Bla{\v{z}} {\v{S}}krlj, Boshko Koloski, and Senja Pollak. 2022.
\newblock Retrieval-efficiency trade-off of unsupervised keyword extraction.
\newblock In \emph{International Conference on Discovery Science}, pages 379--393. Springer.

\bibitem[{Sugandhika and Ahangama(2022)}]{9770051}
Chinthani Sugandhika and Supunmali Ahangama. 2022.
\newblock \href {https://doi.org/10.1109/ACCESS.2022.3172962} {Assessing information quality of wikipedia articles through google’s e-a-t model}.
\newblock \emph{IEEE Access}, 10:52196--52209.

\bibitem[{Thoppilan et~al.(2022)Thoppilan, De~Freitas, Hall, Shazeer, Kulshreshtha, Cheng, Jin, Bos, Baker, Du et~al.}]{thoppilan2022lamda}
Romal Thoppilan, Daniel De~Freitas, Jamie Hall, Noam Shazeer, Apoorv Kulshreshtha, Heng-Tze Cheng, Alicia Jin, Taylor Bos, Leslie Baker, Yu~Du, et~al. 2022.
\newblock Lamda: Language models for dialog applications.
\newblock \emph{arXiv preprint arXiv:2201.08239}.

\bibitem[{Touvron et~al.(2023)Touvron, Lavril, Izacard, Martinet, Lachaux, Lacroix, Rozière, Goyal, Hambro, Azhar, Rodriguez, Joulin, Grave, and Lample}]{touvron2023llamaopenefficientfoundation}
Hugo Touvron, Thibaut Lavril, Gautier Izacard, Xavier Martinet, Marie-Anne Lachaux, Timothée Lacroix, Baptiste Rozière, Naman Goyal, Eric Hambro, Faisal Azhar, Aurelien Rodriguez, Armand Joulin, Edouard Grave, and Guillaume Lample. 2023.
\newblock \href {https://arxiv.org/abs/2302.13971} {Llama: Open and efficient foundation language models}.
\newblock \emph{Preprint}, arXiv:2302.13971.

\bibitem[{Zhang et~al.(2024)Zhang, Yu, Chen, Xiong, Zhu, Qian, Song, Li, Liu, and Li}]{zhang2024retrievalbasedfulllengthwikipediageneration}
Jiebin Zhang, Eugene~J. Yu, Qinyu Chen, Chenhao Xiong, Dawei Zhu, Han Qian, Mingbo Song, Xiaoguang Li, Qun Liu, and Sujian Li. 2024.
\newblock \href {https://arxiv.org/abs/2402.18264} {Retrieval-based full-length wikipedia generation for emergent events}.
\newblock \emph{Preprint}, arXiv:2402.18264.

\end{thebibliography}

\appendix

\clearpage

\appendix
\section*{Appendices}

\section{Details of personal narratives}
\label{appedix:details_personal_arratives}
The details of the personal narratives collected and the corresponding statistics are provided in Table~\ref{tab:data_description}. 

\begin{table*}[!htp]\centering
\scriptsize
\renewcommand{\arraystretch}{1.2}
\resizebox{\textwidth}{!}{
\begin{tabular}{>{\centering\arraybackslash}m{0.1\textwidth}|l|l|l|l|l|l|l|>{\centering\arraybackslash}m{0.08\textwidth}|>{\centering\arraybackslash}m{0.08\textwidth}|>{\centering\arraybackslash}m{0.08\textwidth}
}\toprule
\cellcolor[HTML]{b6d7a8}\textbf{Wikipedia class} &\cellcolor[HTML]{b6d7a8}\textbf{Person} &\cellcolor[HTML]{b6d7a8}\textbf{Gender} &\cellcolor[HTML]{b6d7a8}\textbf{Book name} &\cellcolor[HTML]{b6d7a8}\textbf{Book type} &\cellcolor[HTML]{b6d7a8}\textbf{Author} &\cellcolor[HTML]{b6d7a8}\textbf{Author gender} &\cellcolor[HTML]{b6d7a8}\textbf{wikipedia link} &\cellcolor[HTML]{b6d7a8}\textbf{Number of words} &\cellcolor[HTML]{b6d7a8}\textbf{Number of unique words} &\cellcolor[HTML]{b6d7a8}\textbf{Number of sentences} \\\midrule
\multirow{53}{*}{\textbf{Class B}} &\cellcolor[HTML]{fff2cc}\textbf{John G. B. Adams} &Male &Reminiscences of the Nineteenth Massachusetts regiment &Auto- Biography &Adams, John G. B. &Male &\href{https://en.wikipedia.org/wiki/John\_G.\_B.\_Adams}{wiki/John\_G.\_B.\_Adams} &58258 &10604 &1155 \\
&\cellcolor[HTML]{fff2cc}\textbf{AGA KHAN III} &Male &The Memoirs Of AGA KHAN iii &Auto- Biography &AGHA KHAN iii &Male &\href{https://en.wikipedia.org/wiki/Aga\_Khan\_III}{wiki/Aga\_Khan\_III} &140816 &18809 &3833 \\
&\cellcolor[HTML]{fff2cc}\textbf{Giacinto Achilli} &Male &The imprisonment and deliverance of Dr. Giacinto Achilli &BioGraphy &Eardley, Culling Eardle &Male &\href{https://en.wikipedia.org/wiki/Giacinto\_Achilli}{wiki/Giacinto\_Achilli} &54268 &11535 &772 \\
&\cellcolor[HTML]{fff2cc}\textbf{Hannah Adams} &Female &A memoir of Miss Hannah Adams &Auto- Biography &Adams, Hannah &Female &\href{https://en.wikipedia.org/wiki/Hannah\_Adams}{wiki/Hannah\_Adams} &18729 &5465 &231 \\
&\cellcolor[HTML]{fff2cc}\textbf{John Quincy Adams} &Male &John Quincy Adams &Auto- Biography &John Quincy Adams &Male &\href{https://en.wikipedia.org/wiki/John\_Quincy\_Adams}{wiki/John\_Quincy\_Adams} &9796 &12375 &3677 \\
&\cellcolor[HTML]{fff2cc}\textbf{Halide Edib Adıvar} &Female &Memoirs Of Halide Edib &Auto-Biography &Edib, Halide &Female &\href{https://en.wikipedia.org/wiki/Halide\_Edib\_Ad\%C4\%B1var}{wiki/Halide\_Edib\_Ad\%C4\%B1var} &147334 &14944 &5552 \\
&\cellcolor[HTML]{fff2cc}\textbf{Pope Adrian IV} &Male &Pope Adrian IV, a friend of Ireland, from the Analecta Juris Pontificii &Biograpghy &Chaillot, Louis &Male &\href{https://en.wikipedia.org/wiki/Pope\_Adrian\_IV}{wiki/Pope\_Adrian\_IV} &109165 &15617 &4495 \\
&\cellcolor[HTML]{fff2cc}\textbf{John Abel} &Male &John Jacob Abel, M.D. : investigator, teacher, prophet, 1857-1938 &Auto-BioGraphy &Abel, John Jacob &Male &\href{https://en.wikipedia.org/wiki/John\_Jacob\_Abel}{wiki/John\_Jacob\_Abel} &36263 &7366 &946 \\
&\cellcolor[HTML]{fff2cc}\textbf{Jessie Ackermann} &Female &The world through a woman's eyes &Auto-BioGraphy &Ackermann, Jessie &Female &\href{https://en.wikipedia.org/wiki/Jessie\_Ackermann}{wiki/Jessie\_Ackermann} &70214 &8796 &2820 \\
&\cellcolor[HTML]{fff2cc}\textbf{Adam of Usk} &Male &Chronicon Adae de Usk, A.D. 1377-1421 &BioGraphy &Adam, of Usk, Thompson, Edward Maunde &Male &\href{https://en.wikipedia.org/wiki/Adam\_of\_Usk}{wiki/Adam\_of\_Usk} &140956 &31549 &2607 \\
&\cellcolor[HTML]{fff2cc}\textbf{Robert Walpole} &Male &SIR ROBERT WALPOLE A POLITICAL BIOGRAPHY &BioGraphy &ALEX. CHARLES EWALD &Male &\href{https://en.wikipedia.org/wiki/Robert\_Walpole}{wiki/Robert\_Walpole} &155157 &13391 &8306 \\
&\cellcolor[HTML]{fff2cc}\textbf{Jawaharlal Nehru} &Male &JAWAHARLAL NEHRU An Autobiography &Auto-BioGraphy &JAWAHARLAL NEHRU &Male &\href{https://en.wikipedia.org/wiki/Jawaharlal\_Nehru}{wiki/Jawaharlal\_Nehru} &269112 &15636 &12788 \\
&\cellcolor[HTML]{fff2cc}\textbf{Martin Van Buren} &Male &THE AUTOBIOGRAPHY OF MARTIN VAN BUREN &Auto-BioGraphy &MARTIN VAN BUREN &Male &\href{https://en.wikipedia.org/wiki/Martin\_Van\_Buren}{wiki/Martin\_Van\_Buren} &415430 &17840 &16745 \\
&\cellcolor[HTML]{fff2cc}\textbf{Colonel Sanders} &Male &The Colonel: The Captivating Biography of the Dynamic Founder of a Fast-Food Empire &BioGraphy &John Ed Pearce &Male &\href{https://en.wikipedia.org/wiki/Colonel\_Sanders}{wiki/Colonel\_Sanders} &84743 &8700 &4876 \\
&\cellcolor[HTML]{fff2cc}\textbf{Thomas Paine} &Male &LIFE AND WRITINGS OF THOMAS PAINE &Auto-BioGraphy &THOMAS PAINE &Male &\href{https://en.wikipedia.org/wiki/Thomas\_Paine}{wiki/Thomas\_Paine} &81597 &8396 &2681 \\
&\cellcolor[HTML]{fff2cc}\textbf{Angela Davis} &Female &Angela Davis--an autobiography &Auto-BioGraphy &Angela Davis &Female &\href{https://en.wikipedia.org/wiki/Angela\_Davis}{wiki/Angela\_Davis} &138211 &10403 &7341 \\
&\cellcolor[HTML]{fff2cc}\textbf{H. H. Asquith} &Male &The right hon. H. H. Asquith, M. P. : a biography and appreciation &BioGraphy &Elias, Frank &Male &\href{https://en.wikipedia.org/wiki/H.\_H.\_Asquith}{wiki/H.\_H.\_Asquith} &81433 &8390 &7467 \\
&\cellcolor[HTML]{fff2cc}\textbf{William Makepeace Thackeray} &Male &William Makepeace Thackeray; a biography &BioGraphy &Benjamin, Lewis Saul &Male &\href{https://en.wikipedia.org/wiki/William\_Makepeace\_Thackeray}{wiki/William\_Makepeace\_Thackeray} &99563 &10463 &14841 \\
&\cellcolor[HTML]{fff2cc}\textbf{John Ruskin} &Male &John Ruskin : a bibliographical biography &BioGraphy &Axon, William E. A. &Male &\href{https://en.wikipedia.org/wiki/John\_Ruskin}{wiki/John\_Ruskin} &9415 &2638 &672 \\
&\cellcolor[HTML]{fff2cc}\textbf{Jiddu Krishnamurti} &Male &J. KRISHNAMURTI - A Biography &BioGraphy &Pupul Jayakar &Female &\href{https://en.wikipedia.org/wiki/Jiddu\_Krishnamurti}{wiki/Jiddu\_Krishnamurti} &21221 &2708 &1823 \\
&\cellcolor[HTML]{fff2cc}\textbf{Fatima} &Female &A Brief Biography of Hazrat Fatima &BioGraphy &M.M. Dungersi Ph.D &Male &\href{https://en.wikipedia.org/wiki/Fatima}{wiki/Fatima} &16795 &2182 &1765 \\
&\cellcolor[HTML]{fff2cc}\textbf{Helena Blavatsky} &Female &A Biography of Helena Petrovna Blavatsky &BioGraphy &Howard Murphet &Male &\href{https://en.wikipedia.org/wiki/Helena\_Blavatsky}{wiki/Helena\_Blavatsky} &100260 &10536 &7850 \\
&\cellcolor[HTML]{fff2cc}\textbf{Sheikh Mujibur Rahman} &Male &Mujib: The Architect of Bangla Desh, A Political Biography &BioGraphy &Yatindra Bhatnagar &Male &\href{https://en.wikipedia.org/wiki/Sheikh\_Mujibur\_Rahman}{wiki/Sheikh\_Mujibur\_Rahman} &84636 &8150 &4071 \\
&\cellcolor[HTML]{fff2cc}\textbf{Mullah Omar} &Male &Biography of Mullah Omar &BioGraphy &Taliban group &N/A &\href{https://en.wikipedia.org/wiki/Mullah\_Omar}{/wiki/Mullah\_Omar} &5628 &1490 &232 \\
&\cellcolor[HTML]{fff2cc}\textbf{Guru Tegh Bahadur} &Male &Guru Tegh Bahadur Prophet And Martyr - A Biography &BioGraphy &Dr. Trilochan Singh &Male &\href{https://en.wikipedia.org/wiki/Guru\_Tegh\_Bahadur}{wiki/Guru\_Tegh\_Bahadur} &146122 &15614 &8022 \\
&\cellcolor[HTML]{fff2cc}\textbf{William Cobbett} &Male &WILLIAM COBBETT: A BIOGRAPHY &BioGraphy &EDWARD SMITH &Male &\href{https://en.wikipedia.org/wiki/William\_Cobbett}{wiki/William\_Cobbett} &82537 &9843 &3879 \\
&\cellcolor[HTML]{fff2cc}\textbf{Subhas Chandra Bose} &Male &Subhas Chandra Bose -a Biography &BioGraphy &Gautam Chattopadhyay &Male &\href{https://en.wikipedia.org/wiki/Subhas\_Chandra\_Bose}{wiki/Subhas\_Chandra\_Bose} &43754 &5802 &2245 \\
&\cellcolor[HTML]{fff2cc}\textbf{Sister Nivedita} &Female &THE DEDICATED A BIOGRAPHY OF NIVEDITA &BioGraphy &LIZELLE REYMOND &Female &\href{https://en.wikipedia.org/wiki/Sister\_Nivedita}{wiki/Sister\_Nivedita} &129913 &11665 &7116 \\
&\cellcolor[HTML]{fff2cc}\textbf{Benito Mussolini} &Male &MY AUTOBIOGRAPHY &Auto-BioGraphy &Benito Mussolini &Male &\href{https://en.wikipedia.org/wiki/Benito\_Mussolini}{wiki/Benito\_Mussolini} &90176 &11004 &4987 \\
&\cellcolor[HTML]{fff2cc}\textbf{Orson Welles} &Male &A Biography of Orson Welles &BioGraphy &Frank Brady &Male &\href{https://en.wikipedia.org/wiki/Orson\_Welles}{wiki/Orson\_Welles} &343224 &23388 &15524 \\
&\cellcolor[HTML]{fff2cc}\textbf{Ranjitsinhji} &Male &The biography of Colonel His Highness Shri Sir Ranjitsinhji Vibhaji &BioGraphy &Wild, Roland &Male &\href{https://en.wikipedia.org/wiki/Ranjitsinhji}{wiki/Ranjitsinhji} &107249 &10678 &5536 \\
&\cellcolor[HTML]{fff2cc}\textbf{Abdus Salam} &Male &Abdus Salam A biography &BioGraphy &JAGJIT SINGH &Male &\href{https://en.wikipedia.org/wiki/Abdus\_Salam}{wiki/Abdus\_Salam} &83172 &9968 &3901 \\
&\cellcolor[HTML]{fff2cc}\textbf{Mother Teresa} &Female &Mother Teresa: a biography &BioGraphy &Meg Greene Malvasi &Female &\href{https://en.wikipedia.org/wiki/Mother\_Teresa}{wiki/Mother\_Teresa} &63926 &7696 &3405 \\
&\cellcolor[HTML]{fff2cc}\textbf{Kabir} &Male &Kabir and The Bhagti Movement - Kabir - His Biography - &BioGraphy &Mohan Singh &Male &\href{https://en.wikipedia.org/wiki/Kabir}{wiki/Kabir} &41572 &6821 &3465 \\
&\cellcolor[HTML]{fff2cc}\textbf{Ne Win} &Male &General Ne Win: A Political Biography &BioGraphy &Robert Taylor &Male &\href{https://en.wikipedia.org/wiki/Ne\_Win}{wiki/Ne\_Win} &273760 &16128 &15931 \\
&\cellcolor[HTML]{fff2cc}\textbf{Warren Hastings} &Male &Warren Hastings: a biography &BioGraphy &Trotter, Lionel J. (Lionel James) &Male &\href{https://en.wikipedia.org/wiki/Warren\_Hastings}{wiki/Warren\_Hastings} &98054 &12863 &7049 \\
&\cellcolor[HTML]{fff2cc}\textbf{Florence Nightingale} &Female &Florence Nightingale : a biography &BioGraphy &Willis, Irene Cooper &Female &\href{https://en.wikipedia.org/wiki/Florence\_Nightingale}{wiki/Florence\_Nightingale} &60816 &7516 &2546 \\
&\cellcolor[HTML]{fff2cc}\textbf{Uthman} &Male & The Biography Of Uthman Ibn Affan (R) â Dhun-Noorayn &BioGraphy &Dr. Ali Muhammad Sallaabee &Male &\href{https://en.wikipedia.org/wiki/Uthman}{wiki/Uthman} &205481 &10830 &7946 \\
&\cellcolor[HTML]{fff2cc}\textbf{Golda Meir} &Female &Golda Meir - A Political Biography &BioGraphy &Meron Medzini &Male &\href{https://en.wikipedia.org/wiki/Golda\_Meir}{wiki/Golda\_Meir} &373556 &16289 &18950 \\
&\cellcolor[HTML]{fff2cc}\textbf{Robert Boyle} &Female &Robert Boyle: a biography &BioGraphy &Masson, Flora &Female &\href{https://en.wikipedia.org/wiki/Robert\_Boyle}{wiki/Robert\_Boyle} &97287 &9967 &4153 \\
&\cellcolor[HTML]{fff2cc}\textbf{Annie Besant} &Female &Biography of Annie Besant &BioGraphy &Curuppumullage Jinarajadasa &Male &\href{https://en.wikipedia.org/wiki/Annie\_Besant}{wiki/Annie\_Besant} &6469 &1855 &349 \\
&\cellcolor[HTML]{fff2cc}\textbf{Andrew Carnegie} &Male &Autobiography of Andrew Carnegie &Auto-BioGraphy &Andrew Carnegie &Male &\href{https://en.wikipedia.org/wiki/Andrew\_Carnegie}{wiki/Andrew\_Carnegie} &122002 &10558 &7354 \\
&\cellcolor[HTML]{fff2cc}\textbf{Napoleon} &Male &Napoleon A Biography &BioGraphy &Frank McLynn &Male &\href{https://en.wikipedia.org/wiki/Napoleon}{wiki/Napoleon} &337287 &22717 &14494 \\
&\cellcolor[HTML]{fff2cc}\textbf{Hans Christian Andersen} &Male &Hans Christian Andersen; a biography &BioGraphy &Robert Nisbet Bain &Male &\href{https://en.wikipedia.org/wiki/Hans\_Christian\_Andersen}{wiki/Hans\_Christian\_Andersen} &114414 &14061 &4596 \\
&\cellcolor[HTML]{fff2cc}\textbf{Charles Dickens} &Male &Charles Dickens; a biography from new sources &BioGraphy &Straus, Ralph &Male &\href{https://en.wikipedia.org/wiki/Charles\_Dickens}{wiki/Charles\_Dickens} &108796 &10269 &4944 \\
&\cellcolor[HTML]{fff2cc}\textbf{Alfred Austin} &Male &The autobiography of Alfred Austin &Auto-BioGraphy &Alfred Austin &Male &\href{https://en.wikipedia.org/wiki/Alfred\_Austin}{wiki/Alfred\_Austin} &99556 &13814 &4305 \\
&\cellcolor[HTML]{fff2cc}\textbf{W. G. Grace} &Male &The Memorial biography of Dr. W.G. Grace &BioGraphy &Lord Harris &Male &\href{https://en.wikipedia.org/wiki/W.\_G.\_Grace}{wiki/W.\_G.\_Grace} &131765 &9411 &11727 \\
&\cellcolor[HTML]{fff2cc}\textbf{George Buchanan} &Male &George Buchanan : a biography &BioGraphy & Macmillan, D. (Donald) &Male &\href{https://en.wikipedia.org/wiki/George\_Buchanan}{wiki/George\_Buchanan} &61596 &7750 &2933 \\
&\cellcolor[HTML]{fff2cc}\textbf{Simone de Beauvoir} &Female &Force of circumstance &Auto-biography &Simone de Beauvoir &Female &\href{https://en.wikipedia.org/wiki/Simone\_de\_Beauvoir}{wiki/Simone\_de\_Beauvoir} &305164 &21596 &14959 \\
&\cellcolor[HTML]{fff2cc}\textbf{Sukarno} &Male &SUKARNO: An Autobiography &Auto-BioGraphy &Sukarno &Male &\href{https://en.wikipedia.org/wiki/Sukarno}{wiki/Sukarno} &136268 &13184 &11434 \\
&\cellcolor[HTML]{fff2cc}\textbf{John Keats} &Male &John Keats; a literary biography &BioGraphy &Hancock, Albert Elmer &Male &\href{https://en.wikipedia.org/wiki/John\_Keats}{wiki/John\_Keats} &48693 &8910 &3391 \\
&\cellcolor[HTML]{fff2cc}\textbf{Plato} &Male &Plato: Biography &BioGraphy &Nicolae Sfetcu &Male &\href{https://en.wikipedia.org/wiki/Plato}{wiki/Plato} &5139 &1826 &624 \\
&\cellcolor[HTML]{fff2cc}\textbf{Martin Luther} &Male &Martin Luther King, Jr. : a biography &BioGraphy &Bruns, Roger A. &Male &\href{https://en.wikipedia.org/wiki/Martin\_Luther}{wiki/Martin\_Luther} &63204 &7702 &4243 \\\midrule
\multirow{49}{*}{\textbf{Class C}} &\cellcolor[HTML]{fff2cc}\textbf{John Boyle O'Reilly} &Male &Life of John Boyle O'Reilly : together with his complete poems and speeches &Biography &Roche, James Jeffrey &Male &\href{https://en.wikipedia.org/wiki/John\_Boyle\_O'Reilly}{wiki/John\_Boyle\_O'Reilly} &302285 &23789 &11484 \\
&\cellcolor[HTML]{fff2cc}\textbf{Albert Horsley} &Male &The confessions and autobiography of Harry Orchard &Auto-Biography &Horsley, Albert E &Male &\href{https://en.wikipedia.org/wiki/Albert\_Horsley}{wiki/Albert\_Horsley} &72720 &4850 &2236 \\
&\cellcolor[HTML]{fff2cc}\textbf{Henry Adams} &Male &The education of Henry Adams; an autobiography &Auto-Biography &Henery Adams &Male &\href{https://en.wikipedia.org/wiki/Henry\_Adams}{wiki/Henry\_Adams} &205070 &16000 &6828 \\
&\cellcolor[HTML]{fff2cc}\textbf{Helena Modjeska} &Female &Memories and impressions of Helena Modjeska; an autobiography &Auto-Biography &Helena Modjeska &Female &\href{https://en.wikipedia.org/wiki/Helena\_Modjeska}{wiki/Helena\_Modjeska} &185855 &15289 &7761 \\
&\cellcolor[HTML]{fff2cc}\textbf{Elizabeth Stuart Phelps Ward} &Female &Chapters from a life &Auto-Biography &Elizabeth Stuart Phelps Ward &Female &\href{https://en.wikipedia.org/wiki/Elizabeth\_Stuart\_Phelps\_Ward}{wiki/Elizabeth\_Stuart\_Phelps\_Ward} &73405 &8925 &2710 \\
&\cellcolor[HTML]{fff2cc}\textbf{Robin Bryans} &Male &The Dust Has Never Settled &Auto-Biography &Robin Bryan &Male &\href{https://en.wikipedia.org/wiki/Robin\_Bryans}{wiki/Robin\_Bryans} &342846 &23229 &10436 \\
&\cellcolor[HTML]{fff2cc}\textbf{Henry II of France} &Male &Henry II, King of France 1547-1559 &Biography &Baumgartner, Frederic J &Male &\href{https://en.wikipedia.org/wiki/Henry\_II\_of\_France}{wiki/Henry\_II\_of\_France} &164489 &24162 &4955 \\
&\cellcolor[HTML]{fff2cc}\textbf{Louise Michel} &Female &The Red Virgin: Memoirs Of Louise Michel &Biography &Bullitt Lowry and Elizabeth Ellington Gunter &Male,Female &\href{https://en.wikipedia.org/wiki/Louise\_Michel}{wiki/Louise\_Michel} &108940 &17610 &3779 \\
&\cellcolor[HTML]{fff2cc}\textbf{Jerome } &Male &The life of Saint Jerome : the great doctor of the church : in six books &Biography &Jose de Siguenza, fray &Male &\href{https://en.wikipedia.org/wiki/Jerome}{wiki/Jerome} &221692 &28542 &2430 \\
&\cellcolor[HTML]{fff2cc}\textbf{Joseph O. Shelby} &Male &General Jo Shelby : undefeated rebel &Biography &O'Flaherty, Daniel &Male &\href{https://en.wikipedia.org/wiki/Joseph\_O.\_Shelby}{wiki/Joseph\_O.\_Shelby} &206006 &29298 &5589 \\
&\cellcolor[HTML]{fff2cc}\textbf{Jeanne Guyon} &Female &Autobiography of Madame Guyon &Auto-Biography &Jeanne Guyon &Female &\href{https://en.wikipedia.org/wiki/Jeanne\_Guyon}{wiki/Jeanne\_Guyon} &124764 &18329 &1598 \\
&\cellcolor[HTML]{fff2cc}\textbf{Edwin Austin Abbey} &Male &Edwin Austin Abbey : Royal Academician : the record of his life and work &Biography &Lucas, E. V. (Edward Verrall) &Male &\href{https://en.wikipedia.org/wiki/Edwin\_Austin\_Abbey}{wiki/Edwin\_Austin\_Abbey} &117874 &19434 &4025 \\
&\cellcolor[HTML]{fff2cc}\textbf{Billie Burke} &Female &With a feather on my nose &Auto-Biography &Billie Burke &Female &\href{https://en.wikipedia.org/wiki/Billie\_Burke}{wiki/Billie\_Burke} &72606 &15262 &2739 \\
&\cellcolor[HTML]{fff2cc}\textbf{Brian Halton} &Male &From Coronation Street to a Consummate Chemist &Auto-Biography &Brian Halton &Male &\href{https://en.wikipedia.org/wiki/Brian\_Halton}{wiki/Brian\_Halton} &74728 &9635 &2517 \\
&\cellcolor[HTML]{fff2cc}\textbf{Jean-Jacques Rousseau} &Male &The Confessions of Jean Jacques Rousseau &Biography &Jean-Jacques Rousseau &Male &\href{https://en.wikipedia.org/wiki/Jean-Jacques\_Rousseau}{wiki/Jean-Jacques\_Rousseau} &340071 &18077 &10151 \\
&\cellcolor[HTML]{fff2cc}\textbf{Joanna I of Naples} &Female &The beautiful queen, Joanna I of Naples &Biography &Dale, Darley &Female &\href{https://en.wikipedia.org/wiki/Joanna\_I\_of\_Naples}{wiki/Joanna\_I\_of\_Naples} &88450 &8517 &2138 \\
&\cellcolor[HTML]{fff2cc}\textbf{Kim Jong II} &Male &KIM JONG II BIOGRAPHY &Biography &Foreign Languages Publishing House &N/A &\href{https://en.wikipedia.org/wiki/Kim\_Jong\_Il}{wiki/Kim\_Jong\_Il} &116837 &7869 &4229 \\
&\cellcolor[HTML]{fff2cc}\textbf{David Ferrier} &Male &DAVID FERRIER: A BIOGRAPHY &Biography &JOHN LEYLAND &Male &\href{https://en.wikipedia.org/wiki/David\_Ferrier}{wiki/David\_Ferrier} &3300 &1106 &175 \\
&\cellcolor[HTML]{fff2cc}\textbf{William Henry Harrison} &Male &The life of William Henry Harrison, the people's candidate for the presidency &Biography &Jackson, Isaac R. (Isaac Rand) &Male &\href{https://en.wikipedia.org/wiki/William\_Henry\_Harrison}{wiki/William\_Henry\_Harrison} &48837 &7914 &2842 \\
&\cellcolor[HTML]{fff2cc}\textbf{Cicero} &Male &CICERO A BIOGRAPHY &Biography &TORSTEN PETERSSON &Male &\href{https://en.wikipedia.org/wiki/Cicero}{wiki/Cicero} &250533 &15537 &11773 \\
&\cellcolor[HTML]{fff2cc}\textbf{Thutmose III} &Male &The Military Biography of Egypt's Greatest Warrior King &Biography &RICHARD A. GABRIEL &Male &\href{https://en.wikipedia.org/wiki/Thutmose\_III}{wiki/Thutmose\_III} &91078 &7998 &5173 \\
&\cellcolor[HTML]{fff2cc}\textbf{Edward Gibbon} &Male &AUTOBIOGRAPHY OF EDWARD GIBBON &Auto-Biography &EDWARD GIBBON &Male &\href{https://en.wikipedia.org/wiki/Edward\_Gibbon}{wiki/Edward\_Gibbon} &129290 &14533 &5512 \\
&\cellcolor[HTML]{fff2cc}\textbf{Robert Clive} &Male &CLIVE OF PLASSEY A BIOGRAPHY &Biography &A. MERVYN DAVIES &Male &\href{https://en.wikipedia.org/wiki/Robert\_Clive}{wiki/Robert\_Clive} &214791 &15228 &10049 \\
&\cellcolor[HTML]{fff2cc}\textbf{Alexander Pope} &Male &A POLITICAL BIOGRAPHY OF ALEXANDER POPE &Biography &J. A. Downie &Male &\href{https://en.wikipedia.org/wiki/Alexander\_Pope}{wiki/Alexander\_Pope} &128645 &13289 &6494 \\
&\cellcolor[HTML]{fff2cc}\textbf{O. Henry} &Male &O. HENRY BIOGRAPHY &Biography &C. ALPHONSO SMITH &Male &\href{https://en.wikipedia.org/wiki/O.\_Henry}{wiki/O.\_Henry} &71335 &9520 &4378 \\
&\cellcolor[HTML]{fff2cc}\textbf{Robert Owen} &Male &ROBERT OWEN: A BIOGRAPHY &Biography &FRANK PODMORE &Male &\href{https://en.wikipedia.org/wiki/Robert\_Owen}{wiki/Robert\_Owen} &201973 &14516 &10321 \\
&\cellcolor[HTML]{fff2cc}\textbf{Ayub Khan} &Male &Friends Not Masters A Political Autobiography &Auto-Biography &AYUB KHAN &Male &\href{https://en.wikipedia.org/wiki/Ayub\_Khan}{wiki/Ayub\_Khan} &120448 &9007 &6798 \\
&\cellcolor[HTML]{fff2cc}\textbf{Arthur Balfour} &Male &Mr. Balfour, a biography &Biography &Raymond, E. T., b. &Male &\href{https://en.wikipedia.org/wiki/Arthur\_Balfour}{wiki/Arthur\_Balfour} &68536 &8759 &2802 \\
&\cellcolor[HTML]{fff2cc}\textbf{Ahmad ibn Hanbal} &Male &Ahmed Ibn Hanbal and the Mihna : a biography of the Imam &Biography & Patton, Walter Melville &Male &\href{https://en.wikipedia.org/wiki/Ahmad\_ibn\_Hanbal}{wiki/Ahmad\_ibn\_Hanbal} &54765 &6805 &3723 \\
&\cellcolor[HTML]{fff2cc}\textbf{Oliver Goldsmith} &Male &Oliver Goldsmith : a biography &Biography &Irving, Washington &Male &\href{https://en.wikipedia.org/wiki/Oliver\_Goldsmith}{wiki/Oliver\_Goldsmith} &108293 &11894 &5360 \\
&\cellcolor[HTML]{fff2cc}\textbf{Sarojini Naidu} &Female &Sarojini Naidu: A Biography &Biography &Padmini Sengupta &Female &\href{https://en.wikipedia.org/wiki/Sarojini\_Naidu}{wiki/Sarojini\_Naidu} &135661 &13432 &7557 \\
&\cellcolor[HTML]{fff2cc}\textbf{James Mill} &Male &James Mill. A biography &Biography & Bain, Alexander &Male &\href{https://en.wikipedia.org/wiki/James\_Mill}{wiki/James\_Mill} &174729 &13784 &8377 \\
&\cellcolor[HTML]{fff2cc}\textbf{Paramahansa Yogananda} &Male &Autobiography Of A Yogi &Biography &Paramahansa Yogananda &Male &\href{https://en.wikipedia.org/wiki/Paramahansa\_Yogananda}{wiki/Paramahansa\_Yogananda} &159275 &14495 &11293 \\
&\cellcolor[HTML]{fff2cc}\textbf{Henry Irving} &Male &Sir Henry Irving, a biography &Biography &Percy Hetherington Fitzgerald &Male &\href{https://en.wikipedia.org/wiki/Henry\_Irving}{wiki/Henry\_Irving} &88288 &10122 &4541 \\
&\cellcolor[HTML]{fff2cc}\textbf{Friedrich Engels} &Male &Frederick Engels: A Biography &Biography &Heinrich Gemkow &Male &\href{https://en.wikipedia.org/wiki/Friedrich\_Engels}{wiki/Friedrich\_Engels} &210714 &17074 &10992 \\
&\cellcolor[HTML]{fff2cc}\textbf{Henrik Ibsen} &Male &Henrik Ibsen : a critical biography &Biography &Jaeger, Henrik Bernhard &Male &\href{https://en.wikipedia.org/wiki/Henrik\_Ibsen}{wiki/Henrik\_Ibsen} &73628 &9061 &3539 \\
&\cellcolor[HTML]{fff2cc}\textbf{Bhagat Singh} &Male &Biography Of Bhagat Singh &Biography &M M Juneja &Male &\href{https://en.wikipedia.org/wiki/Bhagat\_Singh}{wiki/Bhagat\_Singh} &64168 &8014 &4966 \\
&\cellcolor[HTML]{fff2cc}\textbf{Helen Keller} &Female &HELEN KELLER - BIOGRAPHY - ENGLISH &Biography &ANNIE SCHRAFF &Female &\href{https://en.wikipedia.org/wiki/Helen\_Keller}{wiki/Helen\_Keller} &7315 &1805 &622 \\
&\cellcolor[HTML]{fff2cc}\textbf{Charles Bradlaugh} &Male &The Biography of Charles Bradlaugh &Biography &Adolphe Headingley &Male &\href{https://en.wikipedia.org/wiki/Charles\_Bradlaugh}{wiki/Charles\_Bradlaugh} &160715 &23756 &8366 \\
&\cellcolor[HTML]{fff2cc}\textbf{Edmund Spenser} &Male &A Biography of Edmund Spenser &Biography &John W. Hales &Male &\href{https://en.wikipedia.org/wiki/Edmund\_Spenser}{wiki/Edmund\_Spenser} &27635 &5471 &1191 \\
&\cellcolor[HTML]{fff2cc}\textbf{William Wordsworth} &Male &William Wordsworth : a biography &Biography &Hood, Edwin Paxton &Male &\href{https://en.wikipedia.org/wiki/William\_Wordsworth}{wiki/William\_Wordsworth} &144566 &19899 &5253 \\
&\cellcolor[HTML]{fff2cc}\textbf{Kim Dae-jung} &Male &THE AUTOBIOGRAPHY OF KIM DAE-JUNG &Auto-Biography &Kim Dae-jung &Male &\href{https://en.wikipedia.org/wiki/Kim\_Dae-jung}{wiki/Kim\_Dae-jung} &400015 &17172 &23786 \\
&\cellcolor[HTML]{fff2cc}\textbf{Ibn Hisham} &Male &The Prophetic Biography" - Sirah Ibn Hisham &Auto-biography &Ibn Hisham &Male &\href{https://en.wikipedia.org/wiki/Ibn\_Hisham}{wiki/Ibn\_Hisham} &342262 &12974 &14664 \\
&\cellcolor[HTML]{fff2cc}\textbf{Giuseppe Garibaldi} &Male &Autobiography Of Giuseppe Garibaldi &Auto-Biography &Giuseppe Garibaldi &Male &\href{https://en.wikipedia.org/wiki/Giuseppe\_Garibaldi}{wiki/Giuseppe\_Garibaldi} &140322 &17312 &5780 \\
&\cellcolor[HTML]{fff2cc}\textbf{Molière} &Male &Moliere, a biography &Biography &Chatfield-Taylor, H. C. (Hobart Chatfield) &Male &\href{https://en.wikipedia.org/wiki/Molière}{wiki/Molière} &134957 &15230 &6020 \\
&\cellcolor[HTML]{fff2cc}\textbf{Timur} &Male &The life of Tamerlane the Great, ... 1653 &Biography &Clarke, Samuel, of Grendon- Underwood, Bucks. &Male &\href{https://en.wikipedia.org/wiki/Timur}{wiki/Timur} &25881 &5857 &774 \\
&\cellcolor[HTML]{fff2cc}\textbf{Satyajit Ray} &Male &SATYAJIT RAY - THE INNER EYE - BIOGRAPHY OF A MASTER FILM MAKER &Biography &Andrew Robinson &Male &\href{https://en.wikipedia.org/wiki/Satyajit\_Ray}{wiki/Satyajit\_Ray} &209956 &16264 &10334 \\
&\cellcolor[HTML]{fff2cc}\textbf{René Descartes} &Male &Biography: René Descartes &Biography &Finkel, B. F. &Male &\href{https://en.wikipedia.org/wiki/René\_Descartes}{wiki/René\_Descartes} &2613 &1006 &113 \\
&\cellcolor[HTML]{fff2cc}\textbf{John Locke} &Male &John Locke : a biography &Biography &Cranston, Maurice &Male &\href{https://en.wikipedia.org/wiki/John\_Locke}{wiki/John\_Locke} &219623 &14834 &14518 \\
\bottomrule
\end{tabular}
}
\caption{\footnotesize Description of the collected writings.}
\label{tab:data_description}
\end{table*}

\section{Prompts}
\label{sec:prompts}
The prompt for standard RAG based approach is represented in Table~\ref{app:prompt_std_RAG}. The prompts for Relevance detection, evidence extraction, evidence verification, and summary generation are represented in Table~\ref{app:prompt_relevant_detection}, Table~\ref{app:prompt_evidence_extraction}, Table~\ref{app:prompt_evidence_verification}, and Table~\ref{app:prompt_summary_generation} respectively.

\section{Baselines details}
\label{appendix:baseline_details}
Since there are no appropriate baselines for this task, we propose two strong baselines along with \rvs{}. 
\noindent\textbf{Key-phrase extraction from personal narrative}: We, first split the personal narrative (autobiography/biography) into several chapters based on the chapter names mentioned. We, then employ three different key-phrase extraction techniques: (i) KeyBert~\cite{grootendorst2020keybert}, (ii) Yake~\cite{campos2020yake} and, (iii) Rakun2~\cite{vskrlj2022retrieval} from each of the chapters to extract 5 key-phrases and take union of these. We vary the number of words $\in \{1, 3\}$ for extracting the key-phrases.\\
\noindent\textbf{Key-phrase focused paragraphs}: Once we extract an initial set of key-phrases, we attempt to generate a relevant and coherent paragraph from the book (i.e., autobiography or biography) related to each of the key-phrases. We employ two different methods for generating key-phrase focused paragraph - 1) Coherence score \cite{jwalapuram-etal-2022-rethinking} based, 2) RAG based.\\
\noindent \textit{1. Coherence score based}: We first split the chapters of the book into a list of sentences using sentence breaks (i.e., ".", "!", "?"). Then we use \texttt{sentence-bert} based embeddings to encode each of the sentences as well as the key-phrase to a 768-dimensional vector space. We measure cosine similarity between the sentences and the key-phrase, and select the top 20 sentences as the initial set ($R$). We first initialize the paragraph ($S$) with the most similar sentence from the $R$. Then for the remaining sentences in $R$, we update $S$ by appending a sentence only if the coherence score\footnote{\url{https://huggingface.co/aisingapore/coherence-momentum}} of the updated $S$ is higher than the actual $S$. We continue this step until we exhaust all the sentences in $R$.\\
\noindent \textit{2. RAG based}: We apply similar retrieval method mentioned in Section~\ref{sec:method}, where we use the key-phrases as query to retrieve top k chunks from the personal narratives. Then we use an LLM to generate a paragraph from the retrieved chunks. 

\if{0}\begin{algorithm}
\caption{Generate relevant and coherent paragraph}
\begin{algorithmic}[1]
\Require{$\text{key-phrase}$: a key-phrase}
\State $\text{R} \gets \text{getTopSentence}(\text{key-phrase})$
\Comment{List of most similar sentences ordered by similarity score}
\State $S \gets \text{R}[0]$
\Comment{Initialize with the most similar sentence}
\For{$i \gets 1$ to $|\text{R}|$}
    \If{$\text{coherence\_score}(S \cup \{\text{R}[i]\}) > \text{coherence\_score}(S)$}
        \State $S \gets S \cup \{\text{R}[i]\}$ \Comment{Add sentence to $S$}
    \EndIf
\EndFor
\State \Return $S$
\end{algorithmic}
\end{algorithm}\fi

\noindent\textbf{Wiki-section to key-phrases map}: Once we obtain the list of important key-phrases ($kp$), and their corresponding key-phrase focused paragraphs ($P$), the next task is to identify the top key-phrases (set to five) among the list of key-phrases that are most relevant to a Wikipedia section pertaining to the personality. We use the \texttt{sentence-bert} to encode the key-phrases, paragraphs, and the Wikipedia sections ($S$). To measure the section-wise similarity to key-phrases, we use three features - cosine similarity between section embeddings and key-phrase embeddings, cosine similarity between section embeddings and the paragraph embeddings, and the cosine similarity between the key-phrase embeddings and paragraph embeddings. We use a weighted score of these 3 features to get the final similarity score between section and a key-phrase.
The weighted similarity between a section $S_i$ and a key-phrase $kp_j$is given by: $\alpha * sim(S_i, kp_j) - \beta * sim(S_i, P_j) + \gamma * sim(kp_j, P_j)$  
 where $\alpha$, $\beta$, and $\gamma$ are hyperparameters. The expression attempts to select those paragraphs ($P_j$) that are similar to the key-phrases but at the same time distant from the section content to avoid inclusion of redundant information.

\begin{table*}[!t]
\tcbset{
  fonttitle=\bfseries,
  boxrule=0.5mm,
  width=\textwidth,
  arc=4mm,
  auto outer arc,
  boxsep=2mm,
}

\begin{tcolorbox}[title=Standard RAG based generation prompt (direct prompting)]
You are an expert in editing Wikipedia biography articles from external resources. You are assigned to expand the content of the given Wikipedia section about the personality: \textbf{"\{person\_name\}"}. You are provided with the section content below which requires expansion:\\

\noindent Section title: \textbf{\{section\_title\}}\\
\noindent Section content: \textbf{\{section\_content\}}\\

Based on the above content, I have gathered some documents below: \\

\noindent Document 1: \textbf{\{chunk1\}}\\
\noindent Document 2: \textbf{\{chunk2\}}\\
\noindent Document 3: \textbf{\{chunk3\}}\\
...\\

As an expert, generate a coherent, insightful and neutral expansion of the "Section content". DO NOT use first person words such as "I", "my". DO NOT use any external information. DO NOT add any duplicate sentence from the "Section content". If it is not possible to expand the content from the documents, say so.
\end{tcolorbox}
\caption{\label{app:prompt_std_RAG} Standard RAG based generation prompt.}
\end{table*}

\begin{table*}[!t]
\tcbset{
  colback=blue!5!white,
  colframe=blue!75!black,
  fonttitle=\bfseries,
  boxrule=0.5mm,
  width=\textwidth,
  arc=4mm,
  auto outer arc,
  boxsep=2mm,
}

\begin{tcolorbox}[title=Relevance detection prompt]
You are an expert in editing Wikipedia biography articles from external resources. You are assigned to expand the content of the given Wikipedia section about the personality: \textbf{"\{person\_name\}"}. You are provided with the section content below which requires expansion:\\

\noindent Section title: \textbf{\{section\_title\}}\\
\noindent Section content: \textbf{\{section\_content\}}\\

Based on the above content, I have gathered some documents below: \\

\noindent Document 1: \textbf{\{chunk1\}}\\
\noindent Document 2: \textbf{\{chunk2\}}\\
\noindent Document 3: \textbf{\{chunk3\}}\\
...\\

As an expert, please identify which document(s) from the list is/are relevant to the above section content. Mention the document ID(s) without any explanation. If you feel no document from the above list is relevant, simply state \textit{"No documents are relevant"}.
\end{tcolorbox}
\caption{\label{app:prompt_relevant_detection} Relevance detection prompt.}
\end{table*}

\begin{table*}
\tcbset{
  fonttitle=\bfseries,
  boxrule=0.5mm,
  width=\textwidth,
  arc=4mm,
  auto outer arc,
  boxsep=2mm,
}

\begin{tcolorbox}[title=Evidence extraction prompt]
\textbf{\{chat history for relevance detection\}}\\

As an expert in Wikipedia editor, can you extract the evidences only from the relevant document(s) you identified, which can be seamlessly integrated with the mentioned section?  Just response the supporting evidences as numbered list without any further details. Format should be - <1. Evidence 1>\textbackslash n<2. Evidence 2>. If you feel that there is no supporting evidence, say \textit{"No evidence."}

\end{tcolorbox}
\caption{\label{app:prompt_evidence_extraction} Evidence extraction prompt.}
\end{table*}

\begin{table*}[!t]
\tcbset{
  fonttitle=\bfseries,
  boxrule=0.5mm,
  width=\textwidth,
  arc=4mm,
  auto outer arc,
  boxsep=2mm,
}

\begin{tcolorbox}[title=Evidence verification prompt]

You are an expert at document reviewing and you are assigned to review whether the given list of evidences are extracted from the below documents\\

\noindent Evidences:\\
\textbf{\{evidences\}}\\

From the above statements can you tell me which statements are actually extracted from the below documents:\\

\noindent Document 1: \textbf{\{chunk1\}}\\
\noindent Document 2: \textbf{\{chunk2\}}\\
\noindent Document 3: \textbf{\{chunk3\}}\\
...\\

Output format should be - <evidence number. evidence>. If there is no evidence extracted from the mentioned documents, say \textit{"None."} 

\end{tcolorbox}
\caption{\label{app:prompt_evidence_verification} Evidence verification prompt.}
\end{table*}

\begin{table*}[!t]
\tcbset{
  fonttitle=\bfseries,
  boxrule=0.5mm,
  width=\textwidth,
  arc=4mm,
  auto outer arc,
  boxsep=2mm,
}

\begin{tcolorbox}[title=Summary generation prompt]

\textbf{\{previous chat history for the evidence collection\}}\\

As an expert in Wikipedia editor, can you make a consize summary from the given evidences, which can be seamlessly integrated with the mentioned section? Make your response as informative as possible without any duplicate information from the original content. Just response the summary without any further details. If you feel that it is not possible to generate a consize summary, say \textit{"Not possible."}\\

Evidences:\\
\textbf{\{evidences\}}
\end{tcolorbox}
\caption{\label{app:prompt_summary_generation} Summary generation prompt.}
\end{table*}

\section{Model implementation details}
\label{appendix:model_implementation}
The retrieval phase employed a maximum marginal relevance (MMR) search with a top-k value set to 4.
For the implementation of \rvs{}, we utilized the \texttt{Llama-3-8b-instruct} model from \textit{HuggingFace}. We set the hyperparameters - max\_new\_tokens: 250, do\_sample:True, temperature:0.7, top\_p:0.9. We set the same set of hyperparameters for each phase of \rvs{}\\ 

\noindent During fine-tuning the same model was fine-tuned on a dataset of 34,576 datapoints, using a learning rate of 2e-5 and a batch size of 16. The training was conducted over 10 epochs, leveraging an NVIDIA A100 GPU with 40 GB memory. \\

\noindent In the baseline, we set $\alpha$ as 3, $\beta$ as 2, and $\gamma$ as 1

\section{Details of calibrated informativeness}
\label{appendix:calibrated_informativeness}
We measure the relative improvement as: $\varDelta Quality = Quality(W_{S} + G_{S}) - Quality(W_{S})$. However, the simple `Informativeness' metric does not take into the account (a) how much new information has been added, and (b) how much appropriate the content is in continuing the existing section. To tackle this, we propose a `Calibrated Informativeness (CI)',  formally defined as: $\varDelta CI = \varDelta Informativeness * \text{Fraction of new added words} \quad * \quad \text{Continuation Score}$ where, the fraction of new added words determines how much new information has been added, and the continuation score determines how much the new content is appropriate in expanding the existing section content. To measure the continuation score we employ a supervised approach by fine-tuning a Llama-3 chat model. We curate a dataset by considering all the \textbf{FA category}\footnote{Note that, we consider B and C category articles during inference.} biographical articles as our training data. Overall 1529 FA category biographies are present in the Wikipedia English corpus. For a given FA page, if there are $n$ paragraphs in a section, we consider the first $(n-1)$ paragraphs as the existing content and consider the $n^\text{th}$ paragraph as the ground truth for generated content. We ignore the section where the number of paragraphs are less than 2. To generate negative examples, we randomly select a paragraph from a Wikipedia section of a different biographical article. Finally, each of the training example would contain an incomplete Wikipedia section (containing $(n-1)$ paragraphs) and an output paragraph ($n^\text{th}$ paragraph for positive case; any random paragraph for negative case). Overall, we have 34,576 datapoints for fine-tuning.\\  Similar to \citet{nogueira-etal-2020-document} we formulate the problem as a binary classification task, and the input prompt is:
\begin{mdframed}[]
\scriptsize
Incomplete content:\quad \{existing content\}\\
Generated content:\quad \{paragraph\}\\
Is the `generated content' an appropriate continuation to the `incomplete content'? Answer yes/no: 
\end{mdframed}
\noindent The model is fine-tuned to produce the words \texttt{yes} or \texttt{no} depending on whether the generated content is an appropriate continuation to the incomplete content. That is, \texttt{yes} and \texttt{no} are the `target words' (i.e., ground truth predictions in the sequence-to-sequence transformation). To generate training and test examples for the models, we iterate over each Wikipedia section
and create (incomplete content, generated content, label) example triples for each positive and negative paragraph. The label is \texttt{yes} if the paragraph is the actual $n^\text{th}$ paragraph for the given incomplete content (positive triple) and \texttt{no} (negative triple) otherwise. 
At inference time, to compute probabilities for each `existing Wikipedia section-generated content' pair, we retrieve the unprocessed next-token probabilities for the tokens \texttt{yes} and \texttt{no}. From these, we calculate the  continuation score as follows. 
\begin{equation}
\scriptsize
\text{Continuation score}_{(W_{S_{i}}, G_{S_{i}})} = \frac{p(yes|P_r)}{p(yes|P_r) + p(no|P_r)}
\end{equation}
where, $W_{S_{i}}$ is the existing Wikipedia section content, $G_{S_{i}}$ is the generated content and $P$ is the prompt. 

\subsection{Effectiveness of calibrated informativeness}
\textcolor{black}{The standard \textit{informativeness} metric focuses solely on the amount of content added, but it does not account for two critical factors: (a) the novelty of the information introduced, and (b) the appropriateness of the new content in relation to the existing section. During manual inspection of qualitative examples, we observed that the simple \textit{informativeness} metric showed significant increases when large amounts of content were generated, regardless of the content’s relevance or quality. To address these shortcomings, we propose a normalized \textit{informativeness} (CI) metric, which incorporates both the novelty of the content and its appropriateness for the section.\\
For instance, in Table~\ref{tab:effectiveness_CI}, for the standard RAG-based approach, the simple \textit{informativeness} score was measured as 27.25, with a new\_word\_ratio of 0.40 and a continuation\_score of 0.45, resulting in a final CI score of 4.97. In contrast, for \rvs{}, the \textit{informativeness} score was 9.86, with a new\_word\_ratio of 0.59 and a continuation\_score of 0.89, yielding a CI score of 5.21. This demonstrates the effectiveness of our proposed \textit{calibrated informativeness} metric, as it provides a more nuanced assessment of both content quality and relevance.}

\begin{table}[H]\centering
\resizebox{\columnwidth}{!}{
\begin{tabular}{l|c|c|c|c}\toprule
\textbf{Approach}& \textbf{Informativeness} & \textbf{New\_word\_ratio}& \textbf{Continuation\_Score} &\textbf{Calibrated Informativeness (CI)} \\
\midrule
Coherence Score-Based &10.17	&0.69	&0.22	&1.55 \\
RAG Paragraph&23.26	&\textbf{0.77}	&0.13	&2.35  \\
Standard RAG-Based &\textbf{27.25}	&0.40	&0.45	&4.97 \\
\rvs{}&9.86	&0.59	&\textbf{0.89}	&\textbf{5.21}\\
\bottomrule
\end{tabular}
}
\caption{\footnotesize Representative example of the effectiveness of \textit{calibrated informativeness}.}\label{tab:effectiveness_CI}
\end{table}



\section{Qualitative examples}
\label{sec:qualitative_examples}
\subsection{Comparative examples of generated content by different methods}
\label{subsec:comparative_examples}
In Table~\ref{tab:comparative_examples} we present a representative example of generated content for a particular Wikipedia section by different approaches.
\begin{table*}[!h]
\caption{\footnotesize Comparison of generated content for each of the methods}
\label{tab:comparative_examples}
\scriptsize
\begin{mdframed}[backgroundcolor=gray!2]
\textbf{Person:  John Quincy Adams}\\ =================================================\\
\textbf{\underline{Existing section: Monroe\_Doctrine}}\\
As the Spanish Empire continued to fracture during Monroe's second term, Adams, Monroe and Clay became increasingly concerned that the "Holy Alliance" of Prussia, Austria, and Russia would seek to bring Spain's erstwhile colonies under their control, to the point of even contemplating a Holy Alliance of their own to defend democracy. In his 1821 Fourth of July address, Adams addressed this issue, noting a shared "chain of sympathy" between the U.S. and Latin America, but arguing for neutrality rather than a Holy Alliance.[83] In 1822, following the conclusion of the Adams–Onís Treaty, the Monroe administration recognized the independence of several Latin American countries, including Argentina and Mexico. In 1823, British Foreign Secretary George Canning suggested that the United States and Britain should work together to preserve the independence of these fledgling republics. The cabinet debated whether to accept the offer, but Adams opposed it. Instead, Adams urged Monroe to publicly declare the United States' opposition to any European attempt to colonize or re-take control of territory in the Americas, while also committing the United States to neutrality in European affairs. In his December 1823 annual message to Congress, Monroe laid out the Monroe Doctrine, which was largely built upon Adams's ideas.[84] In issuing the Monroe Doctrine, the United States displayed a new level of assertiveness in international relations, as the doctrine represented the country's first claim to a sphere of influence. It also marked the country's shift in psychological orientation away from Europe and towards the Americas. Debates over foreign policy would no longer center on relations with Britain and France, but instead focus on western expansion and relations with Native Americans.[85] The doctrine became one of the foundational principles of U.S. foreign policy.[84]\\

\colorbox{gray!40}{\parbox{\textwidth}{\centering \textbf{\rvs{}}}}\\
The Monroe Doctrine was built upon John Quincy Adams' proposal, stating that the American continents should not be considered as subject for future colonization by any European power. The doctrine warned that any interposition by a European power would be viewed as an unfriendly disposition towards the United States, effectively declaring war.

\colorbox{gray!40}{\parbox{\textwidth}{\centering \textbf{Standard RAG based (Direct prompting})}}\\
As the Spanish Empire fractured, Adams, Monroe, and Clay worried the "Holy Alliance" might control former Spanish colonies. Adams, in his 1821 Fourth of July address, emphasized neutrality over a Holy Alliance. Following the Adams–Onís Treaty in 1822, the Monroe administration recognized several Latin American countries' independence. In 1823, Adams opposed a British suggestion to jointly preserve these republics' independence. Instead, he urged Monroe to declare U.S. opposition to European colonization in the Americas while maintaining neutrality in European affairs. Monroe's December 1823 message, influenced by Adams, introduced the Monroe Doctrine, asserting a U.S. sphere of influence and shifting focus from Europe to the Americas. Quincy’s proposal that the Americas were no longer open to European colonization was included verbatim. The doctrine deterred European ambitions, bolstered American confidence, and shifted European powers to prioritize trade over military engagement with the Americas.\\

\colorbox{gray!40}{\parbox{\textwidth}{\centering \textbf{Key-phrase to section mapping (Coherence score based)}}}\\
Monroe was  Under  orders   When the   new   British   minister   demanded   an explanation,  Monroe  replied  angrily  that  American  ships  had  as much  right     the   British   refused,   Congress   declared   British impressment  and  ship  seizures     War  fever  was  infecting  Europe    After  Russia  refused to  cease  trading  with  Britain,  Napol\u00e9on  ordered  French  troops to  the  Russian  border.\\

\colorbox{gray!40}{\parbox{\textwidth}{\centering \textbf{Key-phrase to section mapping (RAG based)}}}\\
John Quincy Adams is best known for his writings on the subject of freedom, and his ideas on the subject have had a profound influence on American thought and policy. His article \"On the Conduct of Foreign Ministers,\" published in 1820, attracted much attention and was seen by many as a blueprint for a successful American foreign policy. Adams's writings on freedom were particularly influential in the years leading up to the Civil War, when many Americans were grappling with questions about the limits of individual freedom and the role of the federal government in regulating society. Adams's ideas helped to shape the way many Americans thought about these issues, and his writings continue to be studied by historians and policymakers today.

\end{mdframed}

\end{table*}

\subsection{Representative example of effectiveness of Relevance detection}
\label{subsec:relevance}
In Table~\ref{tab:example_effectiveness_relevance} we present a representative example to demonstrated how \textit{evidence verification} helps in reducing redundancy.
\begin{table*}[!h]
\caption{\footnotesize A representative example where the \textit{Relevance detection} helps in reducing redundancy.}
\label{tab:example_effectiveness_relevance}
\scriptsize
\begin{mdframed}[backgroundcolor=gray!2]
\textbf{Person:  POPE ADRIAN IV}\\ =================================================\\
\textbf{\underline{Existing section: Death}}\\
At Anagni Hadrian proclaimed the emperor excommunicate and a few days later, to cool himself down [during the hot weather] he started off for a certain fountain along with his attendants. When he got there he drank deeply and at once (according to the story), a fly entered his mouth, stuck to his throat, and could not be shifted by any device of the doctors: and as a result, the pope died.[12]
Burchard of Ursperg's Chronicon Urspergensis, c. 1159By autumn 1159 it may have been clear to Adrian's household and companions that he had not long to live. This may have been at least in part caused by the stresses of his pontificate, suggests Norwich, which although short, was difficult.[267] Pope Adrian died in Anagni[290]—to where he had retired for security against the Emperor[184]—from quinsy[citation needed][note 65] on 1 September 1159. He died, says Norwich, "as many Popes had died before him, an embittered exile; and when death came to him, he welcomed it as a friend".[267] He was buried three days later[4] in an "undistinguished third-century sarcophagus"[267] porphyry tomb of his own choosing.[71][note 66] In 1607, the Italian archaeologist Giovanni Francesco Grimaldi excavated the crypt and in the process opened Adrian's tomb. He described the body, still well preserved, as that of an "undersized man, wearing Turkish slippers on his feet and, on his hand, a ring with a large emerald", and dressed in a dark Chasuble.[267][184]
At the time of Adrian's death, Partner argues, "imperial pressure on the papacy was stronger than it had been since the time of Henry V, and it is not surprising that the cardinals were unable to agree about his successor".[292] It is likely that in the months presaging his death the cardinals were aware of the likelihood of a schism occurring soon afterwards;[143] Freed suggests that thanks to Adrian's own policies, "a split in the College of Cardinals was thus almost preordained", regardless of the Emperor's input.[293] Ullmann suggests that it was the ideological positions of individual cardinals which was shaping—and introducing faction to—the Curia in the last months of Adrian's pontificate.[156] However, Norwich states that Frederick Barbarossa orchestrated the schism himself.[294]
In September 1159—now leading the Emperor's opponents[citation needed]—Adrian had agreed ("but did not swear") to excommunicate Barbarossa.[293] He also did not have time to judge the request of Scottish Legates who had been in Rome since that summer, who were requesting the Diocese of St Andrews be made a metropolitan,[295] and the beatification of Waltheof of Melrose.[296][note 67] One of his final acts was the blessing of his preferred successor, Bernard, Cardinal-Bishop of Porto,[4][note 68] testified Eberhard, Bishop of Bamberg to the Conclave.[157] This, suggests Sayers, could have been Adrian's "masterstroke". The election of Bernard—as a candidate acceptable to the Emperor—may have avoided the future schism.[4] That the Cardinals ended up agreeing with Adrian's choice indicates he had chosen wisely, argues Baumgartner.[94][note 69]
Pope Adrian was buried in St Peter's on 4 September 1159. Present were three Imperial ambassadors who had been in attendance on the Pope when he died. They were Otto of Wittelsbach—who had tried to beat up Cardinal Roland at Besançon—Guido of Biandrate and Heribert of Aachen.[293][note 70] However, as soon as the Emperor heard of the Pope's death, says Madden, he "sent a group of agents and a great deal of money to Rome" in an attempt to secure the election of a successor with pro-Imperial sympathies.[299]\\

\textbf{\underline{Retrieved documents:}}\\
\textit{Document 1:}\\ 
178 DOCUMENTS.

asking the prayers of ‘‘ those who read his book, and those who hear
it read,” he tells us that the news of Pope Adrian’s death had
reached him a little time before, and he adds that his own patron,
Theobald, Archbishop of Canterbury, though still living, was
weighed down by many infirmities.1 Now, Pope Adrian departed
this life in 1159, and the death of Archbishop Theobald happened
in 1161. Elence, Gale and the other editors of John of Salisbury’s
works, without a dissentient voicc, rcfer Metalogicus to the,year 1159.\\
\textit{Document 2:} \\
  Many changes had taken place in the capital of the Christian
world during the two years of his absence. Pope Eugene the
Third had been summoned to his reward, and had had for his
successor the Bishop of Sabina, aged ninety years, who ascended
the Papal Chair under the name of Anastasius the Fourth. On
the 3rd of December, 1154, only a few weeks after Cardinal Break-
speare’s arrival in Rome, the Pontificate of Pope Anastasius was
cut short by death. Rome being in a very disturbed State, the
Cardinals met in St. Peter’s without delay, and with one voice
chose Nicholas Breakspeare as the snecessor of St. Peter to guide
the helm of Holy Church. He at first declined the onerous charge,
but the clergy and laity took up the cry ‘“‘ Nicholas elected by God,”
and at length he bent his shoulders to the burden. He took the
title of Adrian the Fourth, and his coronation was celebrated with
great pomp in St. Peter’s, on the 24th December, 1154.\\
\textit{Document 3:} \\
 this ceremony the Emperor rose and approached for the kiss of
peace. It was now Adrian’s turn. In dignified words he refused
to grant it, and told the Emperor that until the usual homage was
paid in full he would withhold his blessing and refuse to crown
him. Whatever may be our judgment regarding the ceremonial
details of those times, one cannot fail to be struck by the magni-
ficent courage of the Pontiff. The Emperor used every argument
that could be devised to change Adrian’s resolution, but his words
might as well be addressed to the rocks of Sutri. Threats or
entreaties were alike of no avail to move the steady resolution of
the Pope, who next day quitted the camp and returned to Nepi.\\
\textit{Document 4:} \\
 career of Pope Adrian to suppose that such a Pontiff would assign
to such a king the guardianship of the rights and liberties of the
Irish Church. In reply to Father Morris’s line of argument, Miss
Norgate triumphantly appeals to the high opinion entertained
by the English people of the character of their young Angevin
King in the bright morning of his reign, the English Chronicle
attesting that ‘‘ all folk loved him, for he did good justice and
made peace.”’ This however, is not a sufficient reply to the argu-
ment of Father Morris. It is quite true that in the first months
of his reign in 1154, he left nothing undone to ingratiate himself
with the English people, and hence he was for a time idolized by
them, but this did not prevent him from ambitioning at the very
outset of his reign to grasp the rich domains of the Church and
to crush her liberties, and from the letters of the Archbishop of
Canterbury it is more than probable that those designs of Henry\\

\textbf{\underline{Relevant documents identified by LLM (To reduce redundancy):}}\\
\colorbox{green!70}{\parbox{\textwidth}{1, 3}}\\

\textbf{\underline{List of collected evidences:}}\\
1. Pope Adrian IV died in 1159, and his death was known to John of Salisbury, who was writing his book Metalogicus around that time.\\
2. The Pope's death may have been hastened by the stresses of his pontificate, which was marked by difficulties and challenges.\\

\textbf{\underline{Evidence verification:}}\\
1. Pope Adrian IV died in 1159, and his death was known to John of Salisbury, who was writing his book Metalogicus around that time. (Document 1)\\

\textbf{\underline{Generated Summary:}}\\
Pope Adrian IV's death in 1159 was known to John of Salisbury, who wrote his book Metalogicus around that time.

\end{mdframed}

\end{table*}

\subsection{Example of effectiveness of evidence verification}
\label{subsec:example_effectiveness_verification}
In Table~\ref{tab:example_effectiveness_verification} we present a representative example to demonstrated how \textit{evidence verification} helps in reducing redundancy.

\begin{table*}[!h]
\caption{\footnotesize A representative example where the \textit{Evidence verification} helps in reducing duplicate information.}
\label{tab:example_effectiveness_verification}
\scriptsize
\begin{mdframed}[backgroundcolor=gray!2]
\textbf{Person:  Aga Khan III}\\ =================================================\\
\textbf{\underline{Existing section: Early\_life}}\\
He was born in Karachi, Sindh during the British Raj in 1877 (now Pakistan), to Aga Khan II, who migrated from Persia and his third wife,[5] Nawab A'lia Shamsul-Muluk, who was a granddaughter of Fath Ali Shah of Persia. After Eton College, he went on to study at the University of Cambridge.[6]\\

\textbf{\underline{Retrieved documents:}}\\
\textit{Document 1:}\\ 
enough of that. The Aga Khan is descended from the Prophet Mohammed through hisdaughter Fatima and is descended also from the Fatimite Caliphs of Egypt. He is justifiablyproud of his illustrious ancestry. \hl{His grandfather, also known as Aga Khan, by inheritancespiritual head of the Ismailis, was a Persian nobleman, son-in-law of the powerful monarch,Fateh Ali Shah and hereditary chieftain of Kerman.} Smarting under an insult that had beenput upon him he took up arms against a later Shah, Mohammed by name, was worsted andforced to make his escape, attended by a few horsemen, through the deserts of Baluchistan toSind. There he raised a troop of light horse and after various vicissitudes eventually reachedBombay with his two hundred horsemen, his relations, clients and supporters. He acquired avast estate upon which he built palaces, innumerable smaller houses for his dependents andoutbuildings, gardens and fountains. He lived in feudal state and never had less than ahundred horses in his\\
\textit{Document 2:} \\
  it necessary. He has been a great theatergoer; he has loved the opera and the ballet. He is an assiduous reader. He has been occupiedin affairs in which the fate of nations was involved. He has bred horses and raced them. Hehas been on terms of close friendship with kings and princes of the blood royal, maharajahs,viceroys, field marshals, actors and actresses, trainers, golf professionals, society beauties andsociety entertainers. He has founded a university. As head of a widely diffused sect, theIsmailis, he has throughout his life sedulously endeavored to further the welfare, spiritual andmaterial, of his countless followers. Toward the end of this autobiography he remarks that hehas never once been bored. That alone is enough to mark the Aga Khan out as a remarkableman.I must tell the reader at once that I am incompetent to deal with some of his multifariousactivities. I know nothing of racing. I am so little interested in it that one day when I waslunching with the Aga Khan just\\
\textit{Document 3:} \\
 Tehran; others are in Khorassan to the northand east around about Yezd, around Kerman and southward along the coast of the PersianGulf from Bandar Abbas to the borders of Pakistan and Sind, and into Baluchistan. Others arein Afghanistan, in Kabul itself; there are many in Russia and Central Asia, around Yarkand,Kashgar and in many villages and settlements in Sinkiang. In India certain Hindu tribes wereconverted by missionaries sent to them by my ancestor, Shah Islam Shah, and took the nameof Khojas; a similar process of conversion occurred in Burma as recently as the nineteenthcentury.Now that I have brought this brief record of Ismaili origin, vicissitudes and wanderingswithin sight of the contemporary world, it may be timely to give an account in some detail ofthe life and deeds of my grandfather, the first to be known as the Aga Khan, who emergedinto the light of history early in the nineteenth century of the Christian era. His life was (asMr. Justice Arnold observed) "adventurous\\
\textit{Document 4:} \\
 the first to be known as the Aga Khan, who emergedinto the light of history early in the nineteenth century of the Christian era. His life was (asMr. Justice Arnold observed) "adventurous and romantic." He was the hereditary chieftain ofthe important city of Kerman and the son-in-law of the powerful and able Persian monarch,Fateh Ali Shah, holding considerable territorial possessions in addition to his inheritedImamat of the Ismailis.In 1838 he was involved in conflict with the then ruling Emperor Mohammed Shah, forreasons of which Mr. Justice Arnold gave the following account: "Hadji Mirza Ahasi, whohad been the tutor of Mohammed Shah, was during the whole reign of his royal pupil (from1834 to 1848) the Prime Minister of Persia. A Persian of very low origin formerly in theservice of the Aga Khan, had become the chief favorite and minion of the all-powerfulminister. This person, though his patron, had the impudence to demand in marriage for hisson one of the daughters of the Aga Khan,\\

\textbf{\underline{Relevant documents identified by LLM (To reduce redundancy):}}\\
1, 4\\

\textbf{\underline{List of collected evidences:}}\\
1. He was born in Karachi, Sindh during the British Raj in 1877 (now Pakistan), to Aga Khan II, who migrated from Persia and his third wife, Nawab A'lia Shamsul-Muluk, who was a granddaughter of Fath Ali Shah of Persia.\\
2. His grandfather, also known as Aga Khan, by inheritance spiritual head of the Ismailis, was a Persian nobleman, son-in-law of the powerful monarch, Fateh Ali Shah and hereditary chieftain of Kerman.\\

\textbf{\underline{Evidence verification:}}\\
\colorbox{red!70}{\parbox{\textwidth}{1. He was born in Karachi, Sindh during the British Raj in 1877 (now Pakistan), to Aga Khan II, who migrated from Persia and his third wife, Nawab A'lia Shamsul-Muluk, who was a granddaughter of Fath Ali Shah of Persia.}}\\
\colorbox{green!70}{\parbox{\textwidth}{2: His grandfather, also known as Aga Khan, by inheritance spiritual head of the Ismailis, was a Persian nobleman, son-in-law of the powerful monarch, Fateh Ali Shah and hereditary chieftain of Kerman.}} (\colorbox{yellow}{Document 1})\\

\textbf{\underline{Generated Summary:}}\\
Aga Khan III's grandfather, also known as Aga Khan, was a Persian nobleman, son-in-law of Fateh Ali Shah, and hereditary chieftain of Kerman.

\end{mdframed}

\end{table*}

\begin{table*}[!htp]
\centering
\scriptsize
\renewcommand{\arraystretch}{1.1} 
\resizebox{0.7\textwidth}{!}{
\begin{tabular}{
    >{\raggedright\arraybackslash}m{0.08\textwidth}|l|
    >{\centering\arraybackslash}m{0.1\textwidth}|
    >{\centering\arraybackslash}m{0.2\textwidth}|
    >{\centering\arraybackslash}m{0.2\textwidth}|
    >{\centering\arraybackslash}m{0.1\textwidth}
}
\toprule
\textbf{Wikipedia Class} & \textbf{Person} & \textbf{$\varDelta CI$} &\textbf{$\varDelta Understandability$} &\textbf{$\varDelta Readability$} &\textbf{$\varDelta Quality$} \\
\midrule

\multirow{53}{*}{Class B} 
& John\_G.\_B.\_Adams & 5.24 & 0.20 & 0.03 & 1.47 \\
&Aga\_Khan\_III & 59.15 & 0.22 & 0.04 & 15.23 \\
&Giacinto\_Achilli & 33.28 & 0.66 & 0.08 & 8.91 \\
&Hannah\_Adams & 56.07 & 0.49 & 0.06 & 14.61 \\
&John\_Quincy\_Adams & 200.82 & 0.35 & 0.12 & 51.48 \\
&Halide\_Edib\_Adivar & 39.45 & 0.58 & 0.17 & 10.49 \\
&Pope\_Adrian\_IV & 110.26 & 0.14 & 0.02 & 28.21 \\
&John\_Jacob\_Abel & 65.02 & 0.72 & 0.39 & 17.23 \\
&Adam\_of\_Usk & 20.95 & 0.00 & 0.00 & 5.34 \\
&Jessie\_Ackermann & 42.05 & -0.74 & -0.06 & 10.27 \\
&Robert\_Walpole & 17.92 & 0.15 & 0.03 & 4.68 \\
&Jawaharlal\_Nehru & 185.02 & 0.39 & 0.09 & 47.46 \\
&Martin\_Van\_Buren & 77.86 & 0.15 & 0.04 & 19.97 \\
&Colonel\_Sanders & 35.13 & 0.26 & 0.04 & 9.13 \\
&Thomas\_Paine & 44.99 & 0.06 & 0.01 & 11.51 \\
&Angela\_Davis & 48.38 & 0.22 & 0.03 & 12.48 \\
&H.\_H.\_Asquith & 75.21 & 0.05 & 0.01 & 19.21 \\
&William\_Makepeace\_Thackeray & 14.92 & -0.07 & 0.02 & 3.78 \\
&John\_Ruskin & 116.84 & 0.19 & 0.30 & 30.10 \\
&Jiddu\_Krishnamurti & 96.15 & 0.44 & 0.04 & 24.79 \\
&Fatima & 99.38 & 0.39 & 0.06 & 25.60 \\
&Helena\_Blavatsky & 127.13 & -0.04 & 0.02 & 32.41 \\
&Sheikh\_Mujibur\_Rahman & 65.18 & 0.82 & 0.07 & 17.12 \\
&Mullah\_Omar & 58.47 & 0.85 & 0.20 & 15.52 \\
&Guru\_Tegh\_Bahadur & 38.14 & 0.66 & 0.06 & 10.13 \\
&William\_Cobbett & 27.04 & 0.33 & 0.04 & 7.11 \\
&Subhas\_Chandra\_Bose & 58.02 & 2.03 & 0.29 & 16.12 \\
&Sister\_Nivedita & 59.84 & 0.48 & 0.10 & 15.59 \\
&Benito\_Mussolini & 49.00 & 0.05 & -0.03 & 12.51 \\
&Orson\_Welles & 56.40 & 0.50 & 0.10 & 14.73 \\
&Ranjitsinhji & 65.18 & -2.48 & -0.21 & 15.10 \\
&Abdus\_Salam & 42.76 & 0.31 & 0.07 & 11.12 \\
&Mother\_Teresa & 67.50 & 0.59 & 0.39 & 17.79 \\
&Kabir & 42.75 & 0.29 & 0.03 & 11.08 \\
&Ne\_Win & 75.55 & 0.61 & 0.14 & 19.69 \\
&Warren\_Hastings & 57.08 & -0.07 & 0.02 & 14.53 \\
&Florence\_Nightingale & 32.11 & -0.07 & -0.01 & 8.14 \\
&Uthman & 54.07 & -0.38 & 0.30 & 13.77 \\
&Golda\_Meir & 76.72 & 1.21 & 0.15 & 20.33 \\
&Robert\_Boyle & 64.93 & -1.48 & 0.09 & 15.79 \\
&Annie\_Besant & 52.22 & 0.30 & 0.12 & 13.56 \\
&Andrew\_Carnegie & 66.80 & 1.36 & 0.55 & 18.15 \\
&Napoleon & 86.67 & 0.65 & 0.12 & 22.54 \\
&Hans\_Christian\_Andersen & 57.78 & 0.93 & 0.19 & 15.38 \\
&Charles\_Dickens & 53.14 & 0.07 & 0.00 & 13.59 \\
&Alfred\_Austin & 17.56 & 0.13 & 0.10 & 4.61 \\
&W.\_G.\_Grace & 61.19 & -1.57 & -0.20 & 14.60 \\
&George\_Buchanan & 57.10 & 2.11 & 0.63 & 16.15 \\
&Simone\_de\_Beauvoir & 58.01 & 0.04 & 0.03 & 14.83 \\
&Sukarno & 55.70 & 0.60 & 0.08 & 14.59 \\
&John\_Keats & 35.59 & 0.17 & 0.02 & 9.18 \\
&Plato & 39.50 & 0.57 & 0.14 & 10.48 \\
&Martin\_Luther & 44.13 & -0.25 & -0.03 & 11.10 \\
\midrule
&\textbf{Average} & \textbf{61.27} & \textbf{0.27} & \textbf{0.10} & \textbf{15.84} \\
\bottomrule

\multirow{49}{*}{Class C} 
& John\_Boyle\_O'Reilly & 105.19 & 0.48 & 0.08 & 27.14 \\
& Albert\_Horsley & 37.88 & 0.10 & 0.03 & 9.74 \\
& Henry\_Adams & 95.33 & 0.25 & 0.03 & 24.47 \\
& Helena\_Modjeska & 47.30 & 0.44 & 0.06 & 12.35 \\
& Elizabeth\_Stuart\_Phelps\_Ward & 24.49 & 0.34 & 0.12 & 6.51 \\
& Robin\_Bryans & 19.41 & 0.10 & 0.02 & 5.02 \\
& Henry\_II\_of\_France & 58.56 & 0.71 & 0.13 & 15.41 \\
& Louise\_Michel & 49.14 & 0.22 & 0.29 & 12.84 \\
& Jerome & 98.60 & 0.52 & 0.11 & 25.51 \\
& Joseph\_O.\_Shelby & 40.20 & 0.96 & 0.19 & 10.91 \\
& Jeanne\_Guyon & 36.65 & 0.06 & 0.02 & 9.39 \\
& Edwin\_Austin\_Abbey & 21.11 & 0.33 & 0.04 & 5.59 \\
& Billie\_Burke & 46.97 & 0.46 & 0.11 & 12.31 \\
& Brian\_Halton & 17.26 & 0.94 & 0.11 & 5.00 \\
& Jean-Jacques\_Rousseau & 138.89 & 0.05 & 0.01 & 35.45 \\
& Joanna\_I\_of\_Naples & 72.08 & 0.39 & 0.03 & 18.61 \\
&Kim\_Jong\_Il & 70.08 & 1.27 & 0.19 & 18.70 \\
&David\_Ferrier & 7.32 & 0.34 & 0.06 & 2.09 \\
&William\_Henry\_Harrison & 32.17 & 0.67 & 0.09 & 8.63 \\
&Cicero & 40.34 & -0.09 & 0.01 & 10.24 \\
&Thutmose\_III & 24.13 & 0.51 & 0.09 & 6.50 \\
&Edward\_Gibbon & 6.76 & -0.45 & -0.06 & 1.44 \\
&Robert\_Clive & 27.35 & 0.21 & 0.04 & 7.12 \\
&Alexander\_Pope & 20.98 & 0.40 & 0.06 & 5.61 \\
&O.\_Henry & 3.13 & -0.14 & -0.03 & 0.70 \\
&Robert\_Owen & 10.15 & 0.29 & 0.06 & 2.79 \\
&Ayub\_Khan & 60.64 & 1.64 & 0.48 & 16.69 \\
&Arthur\_Balfour & 49.29 & 0.16 & 0.04 & 12.68 \\
&Ahmad\_ibn\_Hanbal & 42.54 & -2.63 & -1.16 & 8.62 \\
&Oliver\_Goldsmith & 39.07 & 0.53 & 0.13 & 10.34 \\
&Sarojini\_Naidu & 107.24 & 1.76 & 0.39 & 28.58 \\
&James\_Mill & 25.65 & 0.35 & 0.10 & 6.80 \\
&Paramahansa\_Yogananda & 76.26 & 0.64 & 0.12 & 19.88 \\
&Henry\_Irving & 47.61 & 0.20 & 0.02 & 12.26 \\
&Friedrich\_Engels & 64.73 & 0.72 & 0.13 & 16.99 \\
&Henrik\_Ibsen & 33.29 & 0.11 & 0.05 & 8.58 \\
&Bhagat\_Singh & 82.96 & 1.09 & 0.13 & 21.84 \\
&Helen\_Keller & 39.27 & 0.13 & 0.02 & 10.10 \\
&Charles\_Bradlaugh & 40.89 & -0.81 & 0.20 & 10.10 \\
&Edmund\_Spenser & 42.04 & 0.23 & 0.06 & 10.89 \\
&William\_Wordsworth & 52.71 & 2.14 & 0.30 & 14.83 \\
&Kim\_Dae-jung & 35.83 & 0.42 & 0.37 & 9.62 \\
&Ibn\_Hisham & 41.81 & -1.00 & -0.21 & 9.97 \\
&Giuseppe\_Garibaldi & 68.64 & 0.16 & 0.05 & 17.63 \\
&Molière & 65.63 & 0.30 & 0.32 & 17.11 \\
&Timur & 31.88 & 0.10 & 0.05 & 8.21 \\
&Satyajit\_Ray & 106.52 & 0.50 & 0.07 & 27.49 \\
&René\_Descartes & 81.83 & 1.11 & 0.19 & 21.61 \\
&John\_Locke & 61.24 & -0.28 & 0.03 & 15.48 \\
\midrule
&\textbf{Average} & \textbf{59.26} & \textbf{0.35} & \textbf{0.08} & \textbf{12.99} \\
\bottomrule
\end{tabular}
}
\caption{\footnotesize Result from the \rvs{} for different Wikipedia biographies using Llama-3-8b-instruct model as LLM}
\label{tab:individual_result_llama-3}
\end{table*}

\section{Additional results}
\subsection{Individual results}
\label{appendix:results}
The results of automatic evaluation of each of the individual personalities is presented in Table~\ref{tab:individual_result_llama-3}.

\subsection{Comparison with other LLMs}
\label{appendix:comparison_llm}
We primarily use \textit{Llama-3-8b-instruct} for our generation tasks. Additionally we conduct the similar generation tasks with few other open-source LLMs to check how they would perform. The average results across different personalities are represented in Table~\ref{tab:comparison_llm}. We observe that the \textit{Llama-3-instruct} significantly (tested using Mann-Whitney U test) outperforms other LLMs in terms of Understandaility and readability.

\begin{table}[!htp]\centering
\scriptsize
\begin{tabular}{lrrrrr}\toprule
\textbf{LLMs} &\textbf{$\varDelta CI$} &\textbf{$\varDelta Und.$} &\textbf{$\varDelta Read.$} &\textbf{$\varDelta Quality$} \\\midrule
Mistral 7B Instruct 0.1 &52.40 &-0.59 &-0.09 &12.92 \\
Llama-2-7B-Instruct &55.04 &-0.12 &0.02 &13.97 \\
Gemma-7B-Instruct &24.26 &-0.37 &-0.09 &5.90 \\
Llama-3-8b-Instruct &60.26 &0.31 &0.09 &14.41 \\
\bottomrule
\end{tabular}
\caption{\footnotesize Performance of other LLMs}
\label{tab:comparison_llm}
\end{table}

\section{Analysis}
\label{sec:additional_analysis}

\subsection{Factual correctness of LLM output}
\label{appendix:factual_correctness}
\textcolor{black}{It is crucial to judge the correctness of the generated content as Wikipedia article should not contain incorrect details. First we designed our 4-step process (relevance detection, evidence extraction, evidence verification, and summarization) specifically to minimize hallucinations and ensure the generated content remains grounded in verified sources. The two critical steps --evidence extraction and evidence verification -- are key to producing factually accurate content.}

In particular, the evidence verification phase is designed to detect and mitigate hallucinations. To achieve this, we separated the chat sessions for this phase (as shown in Figure~\ref{fig:reversum}). During verification, the input to the LLM contains only the ``retrieved chunks'' and ``extracted evidences'' from the source material, with no extraneous information. Furthermore, we instruct the LLM to cite the corresponding chunk number, ensuring that every generated statement is directly grounded in the source.

To assess the correctness, we conducted a qualitative analysis of 50 randomly selected cases where the evidence verification phase yielded results. In this analysis, we did not encounter any instances of hallucinations, underscoring the robustness of our method. In addition, during the summarization phase, the LLM is explicitly instructed to generate content solely from the verified evidence, further reducing the potential for hallucinations. We further conducted a widely used GPT-4 based evaluation for measuring \textit{faithfulness} of the generated summaries relative to the source. The generated content achieved an impressive average \textit{faithfulness score} of 0.95, with all test cases passing—indicating that the content was factually accurate with respect to the source material (i.e., the retrieved chunks). This result provides strong evidence of the reliability and accuracy of our approach. To compute the faithfulness score, we utilized DeepEval \footnote{\url{https://github.com/confident-ai/deepeval}}, a robust tool for evaluating factual consistency in generated text and considered a threshold score of 0.75 as the passing criteria for the individual test cases.

\section{Interface for manual evaluation}
\label{annotation_task_interface}
\begin{figure}[!h]
    \centering
    \includegraphics[width=0.5\textwidth]{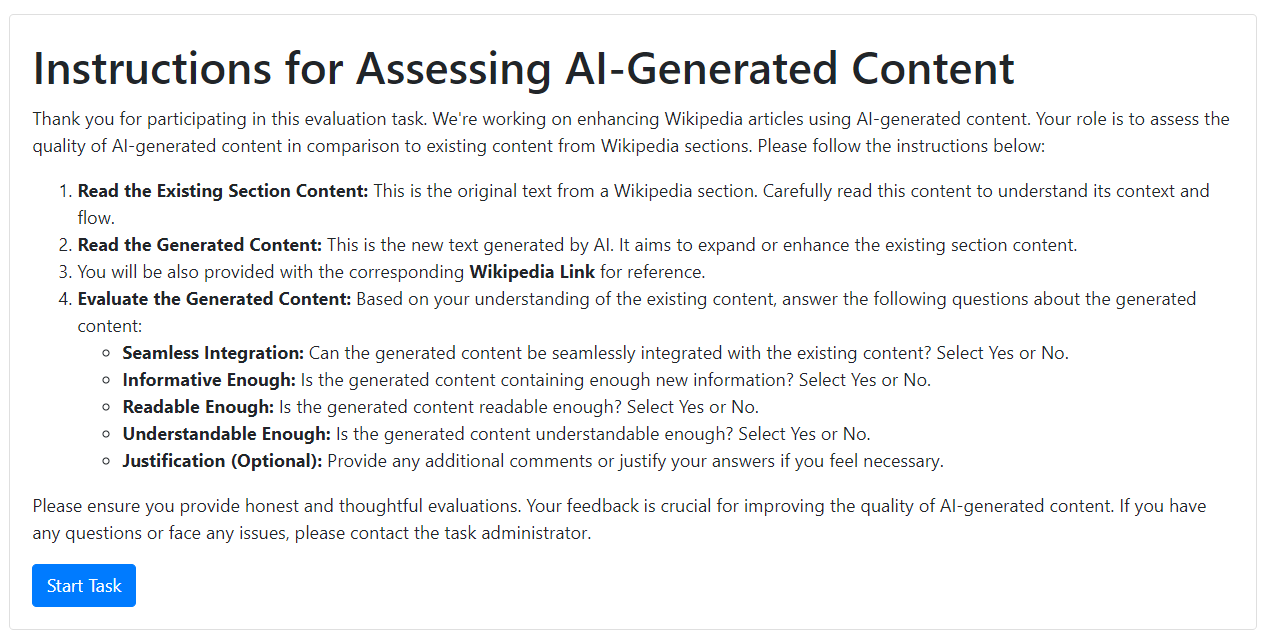}
    \caption{\footnotesize Interface for the annotation task instruction}
    \label{fig:task_instruction}
\end{figure}

\begin{figure}[!h]
    \centering
    \includegraphics[width=0.5\textwidth]{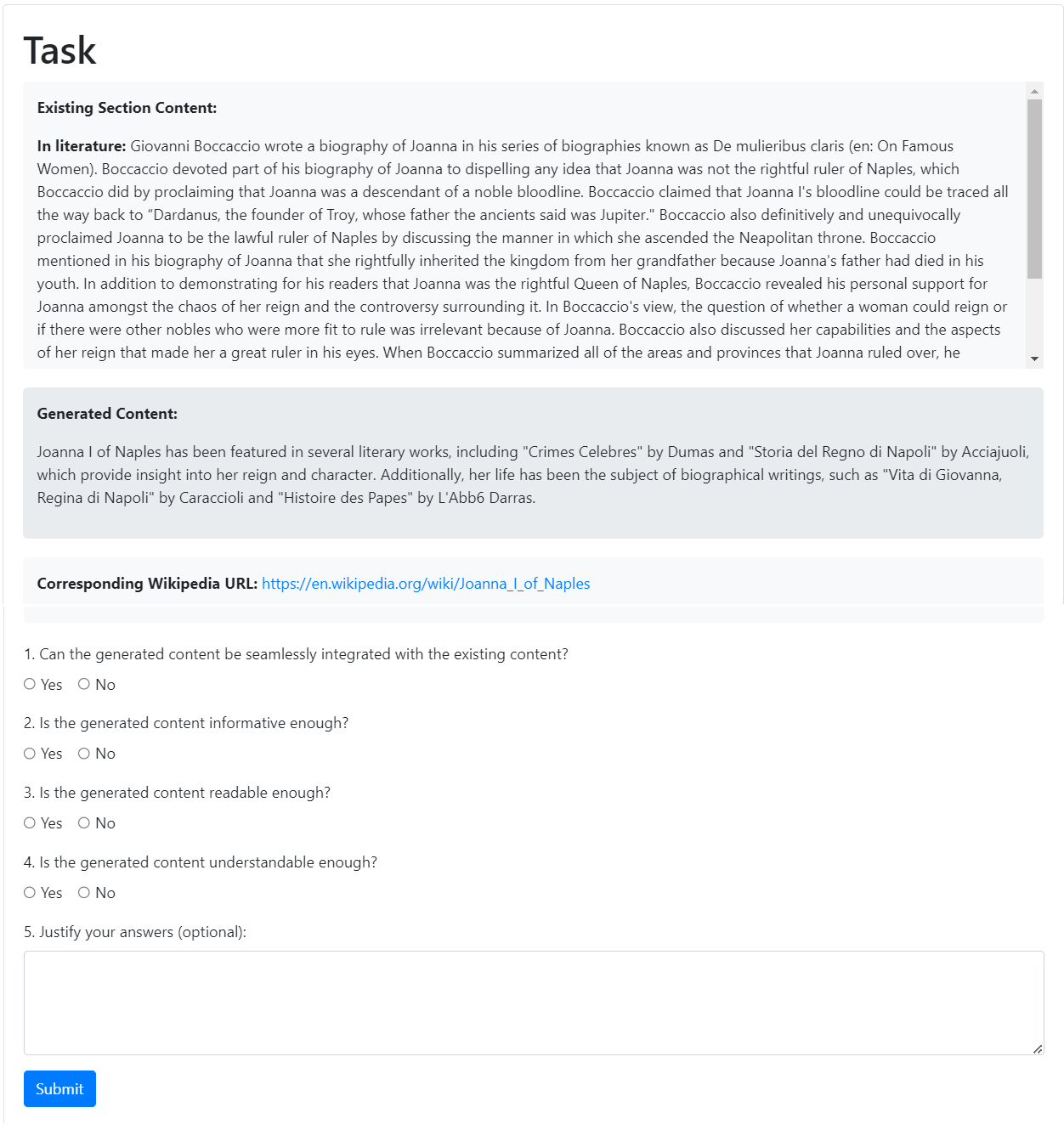}
    \caption{\footnotesize Representative example of an annotation task}
    \label{fig:task_example}
\end{figure}

We prepare a \texttt{Flask} based web interface to manually evaluate the generated content. The task instruction and a representative example for the a task is depicted in Figure~\ref{fig:task_instruction} and Figure~\ref{fig:task_example} respectively.

\end{document}